\PassOptionsToPackage{dvipsnames,table}{xcolor}
\documentclass{article}
\usepackage{iclr2027_conference,times}

\usepackage[utf8]{inputenc}
\usepackage[T1]{fontenc}
\usepackage{hyperref}
\hypersetup{hidelinks}
\usepackage{url}
\usepackage{booktabs}
\usepackage{amsfonts}
\usepackage{amsmath}
\usepackage{amssymb}
\usepackage{amsthm}
\usepackage{nicefrac}
\usepackage{microtype}
\usepackage{xcolor}
\usepackage{graphicx}
\usepackage{multirow}
\usepackage{makecell}
\usepackage{pifont}
\usepackage{enumitem}
\usepackage{caption}
\usepackage{float}
\usepackage{wrapfig}
\usepackage{algorithm}
\usepackage{algpseudocode}

\newcommand{\method}{\textsc{SEER}}

\newtheorem{proposition}{Proposition}[section]
\newtheorem{lemma}[proposition]{Lemma}
\newtheorem{corollary}[proposition]{Corollary}

\newcommand{\maketitlesupplementary}{
  \begin{center}
    \vspace*{-1.6em}
    {\Large\bfseries SEER: Semantic Evidence for Executable Region}
    \\
    {\Large\bfseries Propagation in Open-Vocabulary Semantic Segmentation}\\
    {\Large\bfseries Supplementary Materials}
    \vspace{2.2em}
  \end{center}
}

\title{SEER: Semantic Evidence for Executable\\Region Propagation in Open-Vocabulary\\Semantic Segmentation}

\author{Junyang Chen$^{1,2}$, Xiao Li$^{1,2}$, Yiguo Qiao$^{1,2,*}$, Wenbin Li$^{3}$\\[0.7em]
\normalfont\small $^{1}$School of Computer Science and Engineering,\\
\normalfont\small Southeast University, Nanjing 211189, China\\
\normalfont\small $^{2}$Key Laboratory of Computer Network and Information Integration\\
\normalfont\small (Ministry of Education), Southeast University, Nanjing 211189, China\\
\normalfont\small $^{3}$University of Bath, UK\qquad $^{*}$Corresponding author.}

\iclrfinalcopy
\begin{document}

\maketitle

\vspace{-3mm}
\begin{figure}[ht]
    \centering
    \includegraphics[width=1\linewidth]{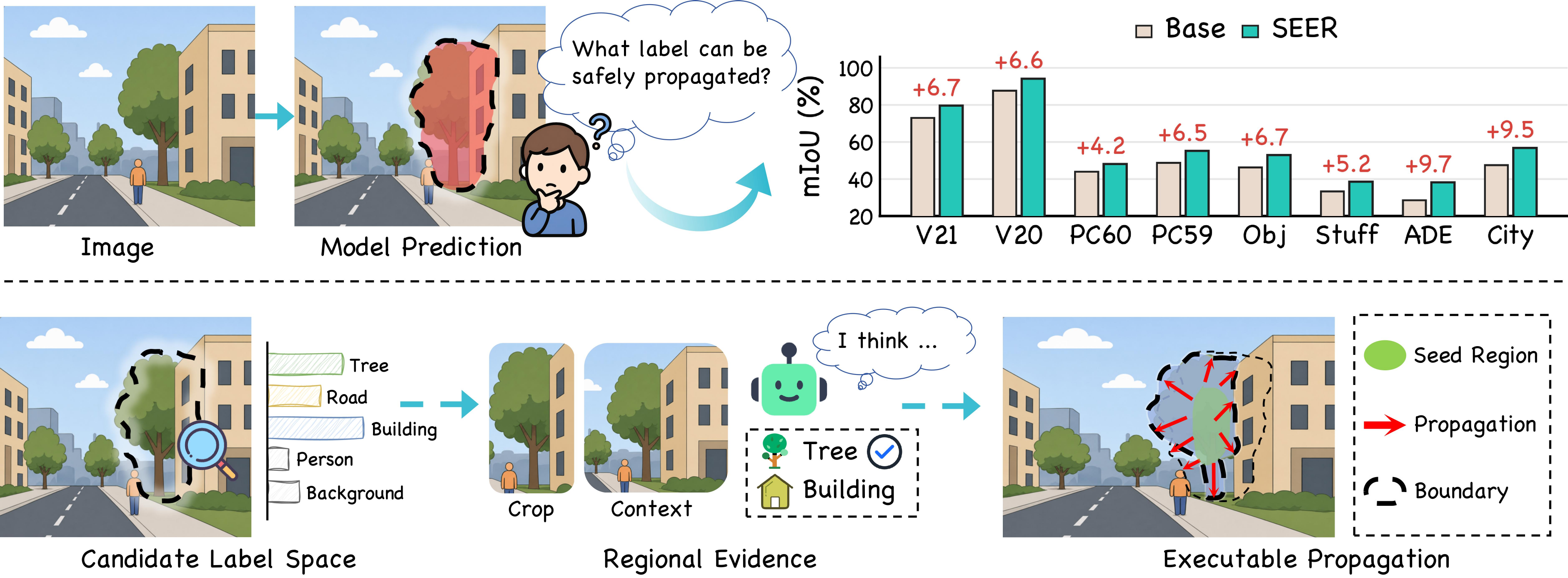}
    % \vspace{-7pt}
    \caption{\textbf{SEER turns regional semantic evidence into executable OVSS label propagation.}
Starting from a fixed OVSS prediction, SEER selects ambiguous proposals, derives compact candidate labels from the active score space, and asks a VLM to verify regional evidence. The OVSS system executes certified evidence: score-supported seeds guide label propagation, while boundary-conservative guards restrict unsafe spreading near object and structure boundaries. The bar chart shows consistent mIoU gains across diverse OVSS benchmarks.}
    \label{fig:teaser}
    \vspace{-1em}
\end{figure}

\begin{abstract}
Training-free open-vocabulary semantic segmentation (OVSS) has made strong progress by adapting CLIP-like representations to dense prediction, but its remaining errors are often local, structured, and tied to the active label space. A successful correction must be visually plausible, legal under the target vocabulary, and executable in the segmentation score field. We propose SEER, a training-free test-time repair layer for Semantic Evidence and Executable Region propagation. SEER casts post-prediction OVSS repair as evidence-conditioned label propagation. The OVSS score space proposes compact candidate labels, a frozen VLM verifies closed-set regional evidence, and an execution gate certifies whether the evidence is legally and semantically actionable. The accepted label is then propagated by an OVSS-side policy: coherent high-confidence regions allow broad propagation, while object- or boundary-risk cases use conservative score-grounded propagation. Across eight standard OVSS benchmarks, SEER reaches 58.0 average mIoU without updating OVSS weights. Matched-input controls isolate the repair mechanism, while resource measurements report the cost of the complete frozen system.
\end{abstract}

\section{Introduction}
\label{sec:intro}
Open-vocabulary semantic segmentation (OVSS) assigns dense labels from an open or user-specified vocabulary. Starting from early open-vocabulary scene parsing~\cite{openvocabscene}, the field has been accelerated by CLIP~\cite{clip} and OpenCLIP~\cite{openclip}. Training-free methods further improve dense prediction by modifying CLIP attention, calibrating patch-text similarity, and injecting visual priors. MaskCLIP~\cite{maskclip} exposes dense CLIP predictions, SCLIP~\cite{sclip} adapts self-attention for segmentation, ProxyCLIP~\cite{proxyclip} introduces proxy features, Trident~\cite{trident} restructures feature aggregation, and CorrCLIP~\cite{corrclip} corrects patch correlation. As these systems become stronger, the bottleneck shifts from global recognition to local residues such as boundary spillovers, object-part confusions, and stuff/object ambiguities.

Residual repair requires more than crop classification. OVSS predictions live under an active label space, where a visually meaningful region can be non-target when its label is absent, while a scene or stuff region becomes executable once the corresponding label enters the vocabulary. A repair action therefore has three coupled requirements. The label must be visually supported, legal under the active vocabulary, and propagated to the right pixels. Regional evidence can suggest a semantic direction, while pixel support must still be recovered from the dense prediction.

General-purpose vision-language models (VLMs) provide a natural source of regional semantic evidence. Qwen-VL~\cite{qwenvl}, LLaVA-OneVision-1.5~\cite{an2025llavaonevision15fullyopenframework}, and GPT-4o~\cite{gpt4o} can compare candidate descriptions and answer localized visual questions. Direct pixel decisions remain unreliable because regional semantics, label-space validity, and pixel execution become entangled in one response. A VLM may name the highlighted region correctly without knowing whether the label is executable in the current OVSS score field or how far it should propagate inside the proposal. We therefore restrict the VLM to regional evidence and retain pixel authority in the segmentation model.

This paper studies how regional VLM evidence can be made executable in OVSS label propagation. Our answer is \method{}, short for \textbf{S}emantic \textbf{E}vidence and \textbf{E}xecutable \textbf{R}egion propagation. \method{} uses prediction-side signals from the baseline mask and score tensor to select auditable proposals, asks a VLM to verify a compact score-derived candidate set with \texttt{uncertain} as an abstention option, and accepts evidence only if it passes executable gates. After certification, the OVSS score field executes the decision through a propagation policy. Coherent high-confidence regions can use broad label propagation, while object-like or boundary-sensitive regions use conservative support derived from the underlying OVSS score field for pixel-level propagation.

This framing keeps the method tied to OVSS. Labels are proposed by open-vocabulary scores, action validity is checked against the active label space, and pixel changes are produced by score-space propagation. Broad propagation is used after closed-set evidence verification, while conservative propagation remains the fallback when boundaries or selector noise make broad support risky. Across eight standard benchmarks, \method{} reaches 58.0 average mIoU with the ViT-L/14 base. Fixed-prediction controls hold the score tensor, proposals, candidates, parser, and audit budget constant while varying the repair signal or executor. We also report complete-system latency and memory, lighter verifiers, proposal-source transfer, sensitivity tests, and residual failures.

This paper makes three contributions.
\begin{itemize}[leftmargin=*,itemsep=2pt,topsep=2pt]
    \item We formulate OVSS post-prediction repair as executable region propagation, separating regional semantic evidence, label-space validity, and pixel support.
    \item \method{} is a training-free evidence-to-execution layer that uses frozen VLM verification while leaving pixel propagation to the OVSS score field.
    \item Systematic diagnostics evaluate compact candidates, executable gates, multi-view evidence, evidence-conditioned propagation, and complete-system cost.
\end{itemize}

\section{Related Work}
\label{sec:related}

\paragraph{Open-vocabulary semantic segmentation.}
OVSS extends zero-shot and open-vocabulary recognition to dense prediction~\cite{openvocabscene,zsseg,spnet}. CLIP~\cite{clip} and OpenCLIP~\cite{openclip} provide the basic image-text alignment, but their image-level training objective does not directly produce spatially coherent patch predictions. Training-free OVSS methods therefore change how CLIP features are extracted, calibrated, or aggregated. MaskCLIP~\cite{maskclip} exposes dense CLIP predictions, CLIP Surgery~\cite{clipsurgery} modifies CLIP behavior for open-vocabulary localization, SCLIP~\cite{sclip} rethinks self-attention for dense inference, CLIPtrase~\cite{cliptrase} studies CLIP's segmentation ability, and ProxyCLIP~\cite{proxyclip} uses proxy attention from stronger visual features. More recent methods improve calibration, residual structure, correlations, and feature aggregation~\cite{lan2024clearclip,naclip,resclip,sc-clip,cliper,cass,dcpclip,trident,corrclip}. These works mainly seek a stronger initial OVSS prediction. \method{} instead repairs residual local errors after that prediction is fixed.

\paragraph{Region priors and label propagation.}
Region and structure priors are widely used in segmentation. Training-based open-vocabulary systems introduce dense supervision, adapters, mask classification, or generative priors~\cite{denseclip,groupvit,ovseg,sideadapter,cat-seg,esc-net,xdecoder,seem,odise,tcl,code}. Class-agnostic and grounding models such as SAM, SAM2, EntitySeg, RegionCLIP, and Grounding DINO provide masks, regions, or localization priors that are useful beyond their original tasks~\cite{sam,sam2,entityseg,regionclip,groundingdino}. Classical CRFs, graph cuts, propagation methods, and modern segmentation architectures also show that spatial consistency and pixel-level confidence matter for reliable masks~\cite{crf,grabcut,deeplabv3plus,segformer,maskformer,mask2former}. \method{} treats each proposal as a candidate domain over which verified evidence may propagate. The propagation distance is conditioned on evidence confidence, region form, score support, and risk.

\paragraph{Vision-language models for visual verification.}
Modern VLMs can answer region-level questions, compare candidate descriptions, and provide visual evidence that is not captured by local score margins~\cite{qwenvl,an2025llavaonevision15fullyopenframework,chen2024internvl,clark2026molmo2,gpt4o,groundingeverything}. This makes them appealing for segmentation repair, but free-form VLM outputs are poorly matched to a fixed OVSS label space. A generic region classifier may name the crop, yet a repair system must also know whether the label is legal and whether the segmentation system can execute the label at pixels. \method{} uses the VLM only as a closed-set verifier over score-derived candidates. The executable decision remains inside the OVSS label and score space, which separates our setting from direct mask correction or ordinary region classification.

\begin{figure*}[t]
    \centering
    \includegraphics[width=1\linewidth]{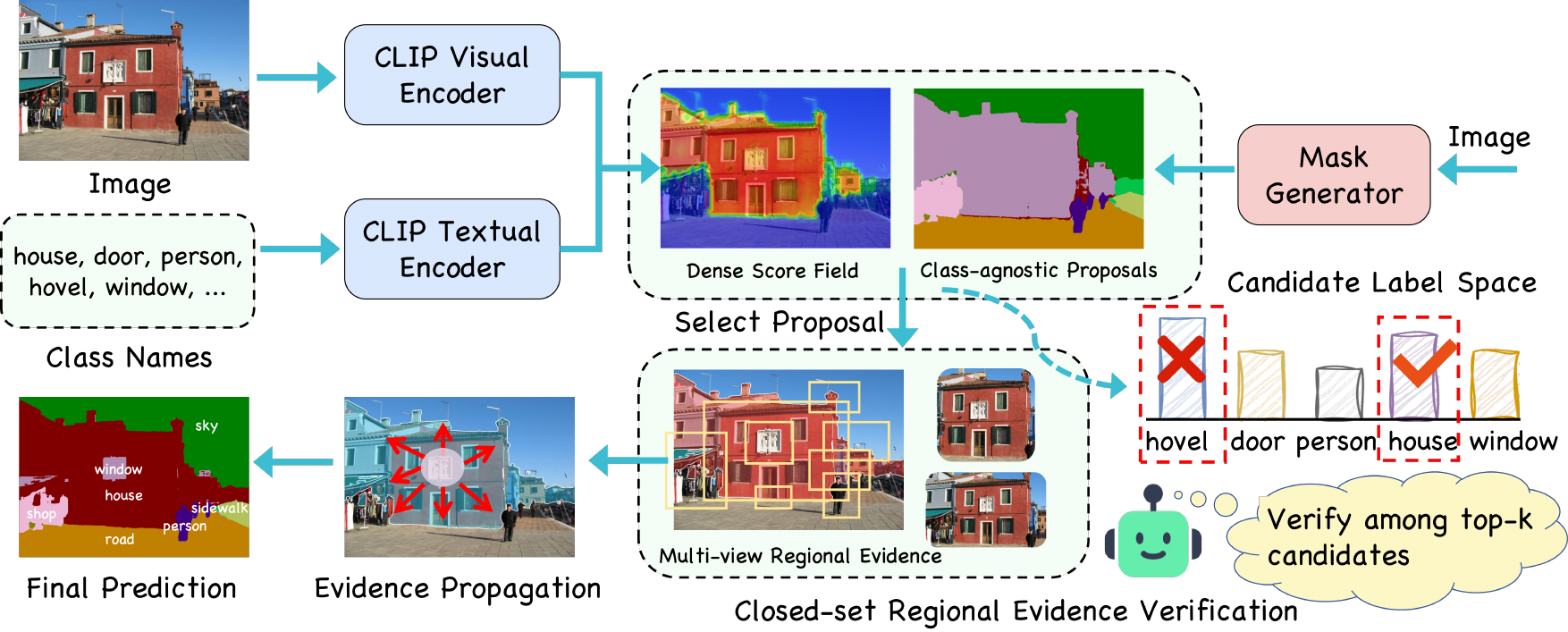}
    \caption{
    \textbf{Overview of \method{}.}
    \method{} converts regional VLM evidence into executable OVSS label propagation. From a fixed prediction, it selects ambiguous proposals, derives score-based candidates, and verifies them with multi-view closed-set reasoning. Score support and vocabulary legality certify the verified label, and the OVSS model controls its pixel propagation.
}
    \label{fig:framework}
    \vspace{-8pt}
\end{figure*}

\section{Method}
\label{sec:method}

\noindent
\textbf{Overview.}
\method{} repairs a fixed OVSS prediction in four steps (Fig.~\ref{fig:framework}). It (1) selects a small proposal subset, (2) builds legal candidate labels from the frozen score tensor, (3) asks a frozen VLM to select one candidate or abstain, and (4) lets the OVSS scores decide which pixels may change. High-confidence coherent regions may use the full proposal; object-like or boundary-sensitive cases use a conservative score-supported subset. The VLM therefore supplies only a regional label decision and never predicts or edits a mask.

\subsection{Problem setup: a label and its pixel support}
Let $I\in\mathbb{R}^{H\times W\times 3}$ be an image and $\Omega=\{1,\ldots,H\}\times\{1,\ldots,W\}$ be the pixel lattice. The active OVSS label space is $\mathcal{C}=\{1,\ldots,C\}$. A fixed OVSS model outputs a prediction map $Y^0\in\mathcal{C}^{H\times W}$ and a dense score tensor $S\in\mathbb{R}^{H\times W\times C}$. We use $S_{p,c}$ for the score of class $c$ at pixel $p$, and $m_p$ for the top-1/top-2 score margin. A fixed SAM2-based class-agnostic proposal bank provides regions $\mathcal{M}=\{M_i\}_{i=1}^{N}$, where $M_i\subseteq\Omega$. The proposal bank is not a semantic predictor in our method; it only defines candidate spatial domains for later evidence verification and score-side execution.

For a proposal $M_i$, a regional label is not yet a repair. It becomes executable only after the system determines how the label should propagate. We write an executable propagation action as
\begin{equation}
    r_i=(\ell_i, W_i),\qquad \ell_i\in\mathcal{C},\quad W_i\subseteq M_i,
\label{eq:repair_action}
\end{equation}
where $\ell_i$ is the verified target label and $W_i$ is the propagation support. Applying $r_i$ updates the current prediction by
\begin{equation}
Y^{t+1}_p=
\begin{cases}
\ell_i, & p\in W_i,\\
Y^{t}_p, & p\notin W_i,
\end{cases}
\label{eq:update}
\end{equation}
where $t$ denotes the execution order. Regional evidence determines the target label, and the OVSS system determines the propagation support $W_i$.

For clarity, we write the three stages of \method{} as
\begin{equation}
\mathcal{M} \xrightarrow{\mathrm{Prioritize}} \mathcal{R}(I),
\qquad
(I,M_i,\mathcal{A}_i) \xrightarrow{\mathrm{Verify}} a_i,
\qquad
(S,Y^0,M_i,a_i) \xrightarrow{\mathrm{Propagate}} (\ell_i,W_i).
\label{eq:seer_factorization}
\end{equation}
Here $\mathcal{R}(I)\subseteq\mathcal{M}$ is the selected auditable subset, $\mathcal{A}_i$ is the candidate label set constructed below, and $a_i$ is the VLM evidence tuple. This factorization is the design principle: region selection limits the VLM budget, verification extracts closed-set evidence, and propagation decides whether the evidence should spread broadly or conservatively.

The pair $(\ell_i,W_i)$ makes the division of labor explicit: the VLM helps choose a regional label, while the OVSS scores choose the pixels that can receive it. A correct label can still increase error if $W_i$ crosses an already-correct boundary, and a safe support can miss useful corrections if it is too small. Appendix~\ref{app:theory_algorithm} gives the exact benefit--harm decomposition and a short non-identifiability result showing why regional evidence alone cannot determine pixel support.

\subsection{Choosing proposals under an audit budget}
\label{sec:proposal_priority}
Querying every proposal is unnecessary and expensive. \method{} selects a budgeted subset using information available before VLM inference. For each proposal, the dominant baseline label is
\begin{equation}
    b_i=\operatorname{mode}\{Y^0_p\mid p\in M_i\}.
\label{eq:base_label}
\end{equation}
The mean score of class $c$ inside the proposal is
\begin{equation}
    \bar{s}_{i,c}=\frac{1}{|M_i|}\sum_{p\in M_i}S_{p,c},
\label{eq:mean_score}
\end{equation}
and the strongest non-baseline competitor is
\begin{equation}
    c_i=\arg\max_{c\in\mathcal{C}\setminus\{b_i\}}\bar{s}_{i,c}.
\label{eq:competitor}
\end{equation}

The priority function is a fixed utility proxy over four prediction-side signals: regional coherence, potential impact, score-space competition, and risk exposure. We write
\begin{equation}
    \pi_i=\pi\!\left(\mathrm{coh}_i,\mathrm{area}_i,\mathrm{comp}_i,\mathrm{risk}_i\right),
\label{eq:priority_function}
\end{equation}
where $\mathrm{coh}_i$ measures the consistency of $Y^0$ inside $M_i$, $\mathrm{area}_i=|M_i|/|\Omega|$, $\mathrm{comp}_i$ measures the score competition between $c_i$ and $b_i$, and $\mathrm{risk}_i$ measures exposure to object or boundary-sensitive predictions. All terms are computed from $Y^0$, $S$, and $M_i$. The exact constants and routing predicates are fixed once and listed in Appendix~\ref{app:implementation}. The priority score is a budget controller that exposes a small auditable set. The selected set under budget $K$ is
\begin{equation}
    \mathcal{R}(I)=\operatorname{TopK}_{M_i\in\mathcal{M}}\pi_i .
\label{eq:topk_regions}
\end{equation}

\subsection{Closed-set regional verification}
\label{sec:semantic_evidence}
For each selected proposal $M_i\in\mathcal{R}(I)$, \method{} constructs a compact candidate set in the OVSS label space:
\begin{equation}
    \mathcal{A}_i=\{b_i\}\cup\mathcal{Q}_i\cup\{\texttt{uncertain}\},
\label{eq:candidate_set}
\end{equation}
where $\mathcal{Q}_i\subseteq\mathcal{C}\setminus\{b_i\}$ contains up to three score-derived alternatives from proposal-averaged scores and local score support. In the default setting, this gives a closed candidate set consisting of the current base label, the top-3 non-base alternatives, and \texttt{uncertain}. The VLM is not asked to search the full vocabulary or invent a label; labels outside $\mathcal{A}_i$ are not executable.

The VLM sees multiple views of the same proposal, including the full image with $M_i$ highlighted and cropped views around the proposal. It returns an evidence tuple
\begin{equation}
    a_i=(\hat{\ell}_i,\mathrm{conf}_i,\mathrm{form}_i),
\label{eq:audit_tuple}
\end{equation}
where $\hat{\ell}_i\in\mathcal{A}_i$ is the selected label, $\mathrm{conf}_i\in\{\texttt{high},\texttt{medium},\texttt{low}\}$ is confidence, and $\mathrm{form}_i$ indicates whether the region is coherent, boundary-like, mixed, or unclear. This tuple specifies a label direction and a region form; pixel support is still decided by the OVSS score field.

\subsection{Filtering VLM answers before writeback}
\label{sec:evidence_gate}
The VLM answer is filtered before it can affect any pixel. The executable indicator is
\begin{equation}
 e_i =
\mathbf{1}\!\left[
\hat{\ell}_i\in\mathcal{C},\;
\hat{\ell}_i\neq b_i,\;
\hat{\ell}_i\neq\texttt{uncertain},\;
\mathrm{conf}_i=\texttt{high},\;
\mathrm{form}_i \notin \{\mathrm{mixed}, \mathrm{unclear}\}
\right].
\label{eq:execution_gate}
\end{equation}
If $e_i=0$, the proposal leaves the prediction unchanged. If $e_i=1$, the target label is $\ell_i=\hat{\ell}_i$. Thus every accepted answer must be legal, high-confidence, different from the current label, and not mixed or unclear. Boundary-like answers may still be used, but only through conservative pixel support. Section~\ref{sec:mechanism_analysis} and Appendix~\ref{app:gate_efficiency} test the filter and propagation choices under matched inputs.

\subsection{Score-controlled pixel updates}
\label{sec:pixel_execution}
Once the label is certified, the remaining question is how far the evidence should propagate. \method{} uses a policy family $\Gamma$ that maps certified evidence and OVSS scores to a support set:
\begin{equation}
    W_i = \Gamma(S,Y^0,M_i,\ell_i,a_i),\qquad W_i\subseteq M_i.
\label{eq:prop_policy}
\end{equation}
We use three propagation regimes. \textit{Coherent propagation} sets $W_i=M_i$ when a region is coherent, high-confidence, and low-risk; in this case, the proposal is treated as a semantic unit. \textit{Score-grounded propagation} propagates only to pixels where the target remains supported by the OVSS score field. \textit{Boundary-conservative propagation} restricts propagation to low-margin pixels for object-like or risk-sensitive cases.

For score-grounded propagation, let
\begin{equation}
    y^{\mathrm{top}}_p=\arg\max_{c\in\mathcal{C}} S_{p,c}.
\label{eq:top_label}
\end{equation}
We use the support predicate
\begin{equation}
\rho(p,\ell_i,b_i)=
\mathbf{1}\!\left[
 y^{\mathrm{top}}_p=\ell_i
 \;\vee\;
 S_{p,\ell_i}\ge S_{p,b_i}-\tau_s
 \;\vee\;
 \big(m_p\le\tau_m \wedge S_{p,\ell_i}\ge S_{p,b_i}-\tau_l\big)
\right],
\label{eq:score_support}
\end{equation}
where $\tau_s,\tau_m,\tau_l$ are fixed thresholds. The score-grounded support is $W_i=\{p\in M_i\mid e_i=1,\rho(p,\ell_i,b_i)=1\}$; the boundary-conservative support further restricts it to low-margin pixels. In practice, \method{} uses the same deterministic contract for all datasets: broad propagation is reserved for reliable context/stuff evidence, while object, small-region, boundary, or high-risk cases use conservative support after closed-set certification.

\section{Experiments}
\label{sec:experiments}

\subsection{Experimental Setup}
\label{sec:exp_setup}
\noindent\textbf{Evaluation protocol.}
Following common open-vocabulary semantic segmentation (OVSS) protocols~\cite{maskclip,sclip,proxyclip,trident,corrclip}, we evaluate on eight benchmarks from six validation sets: PASCAL VOC with and without background (VOC21/VOC20)~\cite{voc12}, PASCAL Context with and without background (PC60/PC59)~\cite{pascalcontext}, COCO Object and COCO-Stuff164K~\cite{coco}, ADE20K~\cite{ade20k}, and Cityscapes~\cite{cityscapes}. We report mean Intersection over Union (mIoU). Comparisons with prior methods follow the CorrCLIP-style evaluation configuration: published numbers are grouped by CLIP backbone size, and our method is evaluated with both CLIP ViT-B/16 and ViT-L/14 settings.

\noindent\textbf{Implementation details.}
We follow the CorrCLIP evaluation configuration, while the repair input itself is a fixed CLIP score field $S$ and its induced prediction $Y^0$. Region domains come from a fixed SAM2~\citep{sam2} proposal bank and are never treated as semantic predictions. The default verifier is the locally deployable Qwen3.5-9B~\citep{qwenvl}, chosen for strong regional recognition, open weights, and deterministic decoding; it is not fine-tuned on any evaluation benchmark. Its public documentation does not provide a sample-level pretraining manifest, so we do not assume zero overlap with segmentation or grounding data and instead test replacement by three other VLM families and CLIP. The VLM receives proposal views and legal label names. Ground truth, target masks, and pixel-write supervision are excluded.

For mechanism studies, we fix $Y^0$, $S$, the proposal bank, candidates, parser, gate, propagation policy, audit inputs, and evaluation split; only the tested repair signal or executor changes. This \textit{fixed-prediction protocol} uses MetaCLIP FullCC ViT-B/16 QuickGELU, frozen DINO ViT-B/8, benchmark-native mappings, and aspect-ratio-preserving resolution $(2048,336)$ for VOC/PC and $(2048,448)$ for City/ADE. Its initial scores are a separate locally frozen diagnostic state, not a reproduction of the published CorrCLIP row in Tab.~\ref{tab:main_results}. The constants were selected once from aggregate diagnostics on the evaluated natural-image validation domains, then frozen across all eight benchmarks, two backbones, four OVSS bases, and every verifier; we do not claim domain-blind development or medical/remote-sensing validation. Appendix~\ref{app:implementation} lists the complete contract and sensitivity tests.

\subsection{Comparison with State-of-the-Art Methods}
\label{sec:main_results}
\begin{table*}[t]
\centering
\small
\setlength{\tabcolsep}{3pt}
\caption{\textbf{Complete-system accuracy on eight OVSS benchmarks.} Published baseline numbers follow the CorrCLIP-style evaluation configuration and are grouped by CLIP size. \method{} is an accuracy--cost point, not an equal-resource comparison: it adds fixed SAM2 proposals and sparse local Qwen3.5-9B verification to CorrCLIP. Paired end-to-end cost and matched-input attribution are reported in Tab.~\ref{tab:efficiency} and Tab.~\ref{tab:fair_nonvlm_main}.}
\label{tab:main_results}
\vspace{-2mm}
\resizebox{\textwidth}{!}{
\begin{tabular}{lc|llllllll|l}
\toprule
Method & Size & VOC21 & VOC20 & PC60 & PC59 & Object & Stuff & ADE & City & Avg. \\
\midrule
\rowcolor{ForestGreen!8}
\multicolumn{11}{c}{\textbf{\emph{Training-based}}} \\
TCL~\cite{tcl} & \multirow{5}{*}{ViT-B/16} & 55.0 & 83.2 & 30.4 & 33.9 & 31.6 & 22.4 & 17.1 & 24.0 & 37.2 \\
CLIP-DINOiser~\cite{clip-dinoier} & & 62.1 & 80.9 & 32.4 & 35.9 & 34.8 & 24.6 & 20.0 & 31.7 & 40.3 \\
CoDe~\cite{code} & & 57.7 & -- & 30.5 & -- & 32.3 & 23.9 & 17.7 & 28.9 & -- \\
LoGoSeg~\cite{logoseg} & & 78.6 & 95.4 & -- & 58.2 & -- & -- & 32.6 & -- & -- \\
ESC-Net~\cite{esc-net} & & 80.1 & 97.3 & -- & 59.0 & -- & -- & 35.6 & -- & -- \\
\midrule
\rowcolor{ForestGreen!8}
\multicolumn{11}{c}{\textbf{\emph{Training-free}}} \\
CLIP~\cite{clip} & \multirow{15}{*}{ViT-B/16} & 11.5 & 41.9 & 4.4 & 9.2 & 1.6 & 4.4 & 2.9 & 5.0 & 10.1 \\
MaskCLIP~\cite{maskclip} & & 38.8 & 74.9 & 23.6 & 26.4 & 20.6 & 16.4 & 9.8 & 12.6 & 27.9 \\
ClearCLIP~\cite{lan2024clearclip} & & 51.8 & 80.9 & 32.6 & 35.9 & 33.0 & 23.9 & 16.7 & 30.0 & 38.1 \\
SCLIP~\cite{sclip} & & 59.1 & 80.4 & 30.4 & 34.2 & 30.5 & 22.4 & 16.1 & 32.2 & 38.2 \\
ProxyCLIP~\cite{proxyclip} & & 61.3 & 80.3 & 35.3 & 39.1 & 37.5 & 26.5 & 20.2 & 38.1 & 42.3 \\
LaVG~\cite{lavg} & & 62.1 & 82.5 & 31.6 & 34.7 & 34.2 & 23.2 & 15.8 & 26.2 & 38.8 \\
CLIPtrase~\cite{cliptrase} & & 53.0 & 81.2 & 30.8 & 34.9 & \underline{44.8} & 24.1 & 17.0 & -- & -- \\
NACLIP~\cite{naclip} & & 64.1 & 83.0 & 35.0 & 38.4 & 36.2 & 25.7 & 19.1 & 38.3 & 42.5 \\
Trident~\cite{trident} & & 67.1 & 84.5 & 38.6 & 42.2 & 41.1 & 28.3 & 21.9 & 42.9 & 45.8 \\
ResCLIP~\cite{resclip} & & 61.1 & 86.0 & 33.5 & 36.8 & 35.0 & 24.7 & 18.0 & 35.9 & 41.4 \\
SC-CLIP~\cite{sc-clip} & & 64.6 & 84.3 & 36.8 & 40.1 & 37.7 & 26.6 & 20.1 & 41.0 & 43.9 \\
CLIPer~\cite{cliper} & & 65.9 & 85.2 & 37.6 & 41.7 & 39.0 & 27.5 & 21.4 & -- & -- \\
CASS~\cite{cass} & & 65.8 & 87.8 & 36.7 & 40.2 & 37.8 & 26.7 & 20.4 & 39.4 & 44.4 \\
CorrCLIP~\cite{corrclip} & & \underline{74.8} & \underline{88.8} & \underline{44.2} & \underline{48.8} & 43.7 & \underline{31.6} & \underline{26.9} & \underline{49.4} & \underline{51.0} \\
\rowcolor{SkyBlue!16}
\method{} (Ours) & & \textbf{79.3} \textcolor{blue}{(+4.5)} & \textbf{93.8} \textcolor{blue}{(+5.0)} & \textbf{47.7} \textcolor{blue}{(+3.5)} & \textbf{53.9} \textcolor{blue}{(+5.1)} & \textbf{48.5} \textcolor{blue}{(+3.7)} & \textbf{36.2} \textcolor{blue}{(+4.6)} & \textbf{35.2} \textcolor{blue}{(+8.3)} & \textbf{55.0} \textcolor{blue}{(+5.6)} & \textbf{56.2} \textcolor{blue}{(+5.2)} \\
\midrule
FreeDA~\cite{freeda} & \multirow{8}{*}{ViT-L/14} & 55.4 & 87.9 & 38.3 & 43.5 & 37.4 & 28.8 & 23.2 & 36.7 & 43.9 \\
CaR~\cite{car} & & 67.6 & 91.4 & 30.5 & 39.5 & 36.6 & -- & 17.7 & -- & -- \\
ProxyCLIP~\cite{proxyclip} & & 60.6 & 83.2 & 34.5 & 37.7 & 39.2 & 25.6 & 22.6 & 40.1 & 43.0 \\
ResCLIP~\cite{resclip} & & 54.1 & 85.5 & 30.9 & 34.5 & 32.5 & 23.4 & 18.2 & 33.7 & 39.1 \\
SC-CLIP~\cite{sc-clip} & & 65.0 & 88.3 & 36.9 & 40.6 & 40.5 & 26.9 & 21.7 & 41.3 & 45.2 \\
CLIPer~\cite{cliper} & & 69.8 & 90.0 & 38.0 & 43.6 & 43.3 & 28.7 & 24.4 & -- & -- \\
CorrCLIP~\cite{corrclip} & & \underline{76.7} & \underline{91.5} & \underline{44.9} & \underline{50.8} & \underline{49.4} & \underline{34.0} & \underline{30.7} & \underline{51.1} & \underline{53.6} \\
\rowcolor{SkyBlue!16}
\method{} (Ours) & & \textbf{79.6} \textcolor{blue}{(+2.9)} & \textbf{94.2} \textcolor{blue}{(+2.7)} & \textbf{48.2} \textcolor{blue}{(+3.3)} & \textbf{55.2} \textcolor{blue}{(+4.4)} & \textbf{52.9} \textcolor{blue}{(+3.5)} & \textbf{38.6} \textcolor{blue}{(+4.6)} & \textbf{38.3} \textcolor{blue}{(+7.6)} & \textbf{56.9} \textcolor{blue}{(+5.8)} & \textbf{58.0} \textcolor{blue}{(+4.4)} \\
\bottomrule
\end{tabular}}
\vspace{-3mm}
\end{table*}

\paragraph{Complete-system comparison.}
Table~\ref{tab:main_results} compares complete-system accuracy using published baseline results; it does not imply matched hardware or equal external resources. With both ViT-B/16 and ViT-L/14, SEER improves its corresponding CorrCLIP base on every benchmark, with the largest gains on ADE20K and Cityscapes. The remaining experiments use one frozen prediction state and matched proposal/audit inputs to attribute the gain, while Tab.~\ref{tab:efficiency} reports the paired cost on our hardware.

\subsection{Fixed-Prediction Repair Analysis}
\label{sec:fixed_base}
\paragraph{Fixed-prediction repair.}
The fixed-prediction diagnostic in Fig.~\ref{fig:teaser} reports the controlled setting used in the mechanism analysis. The prediction map and score tensor are fixed, and only the evidence-to-execution layer is applied. Gains across eight benchmarks rule out changes in backbone, resolution, or training recipe, supporting the premise that residual OVSS errors are repairable once regional evidence becomes executable.

\paragraph{Plug-in transfer across OVSS bases.}
We next ask whether the repair layer is tied to one score geometry. The same SEER pipeline is applied to ProxyCLIP, Trident, ResCLIP, and CorrCLIP: proposals, candidates, VLM auditing, filters, propagation rules, coefficients, and thresholds stay fixed across bases and benchmarks. Only the OVSS prediction and score cache change. Every reported base--benchmark pair improves, with mean gains from 4.53 to 5.93 mIoU; full results and confidence intervals appear in Appendix~\ref{app:cross_base}.

\subsection{Mechanism Analysis}
\label{sec:mechanism_analysis}
\paragraph{Component chain.}
The next experiments follow the method decomposition. The full-chain diagnostic in Table~\ref{tab:component_summary_main} compares SEER with simpler repair configurations. Each alternative breaks one link in the deterministic execution contract: visual verification, label-space restriction, multi-view context, or high-recall coherent propagation. Within this diagnostic, all variants reduce the average gain, supporting SEER as a coupled design whose gain depends on both regional evidence and score-side execution.

\begin{table}[t]
\centering
\small
\setlength{\tabcolsep}{6pt}
\caption{\textbf{Component diagnostics.} Each row is an alternative repair setting, not a cumulative build-up. The initial values are an independently frozen diagnostic $S,Y^0$, not the published CorrCLIP row in Tab.~\ref{tab:main_results}; every delta uses this identical state.}
\label{tab:component_summary_main}
\vspace{1mm}
\resizebox{0.98\linewidth}{!}{
\begin{tabular}{lcccc|rrrrr}
\toprule
Setting & VLM & Closed & Multi-view & Coherent & VOC21 & PC60 & City & ADE & Avg. \\
\midrule
Initial prediction &  &  &  &  & 72.88 & 44.01 & 47.41 & 28.63 & 48.23 \\
Score-only rule &  &  &  & \checkmark & +2.10 & +3.80 & +2.53 & +2.13 & +2.64 \\
Free-form VLM & \checkmark &  & \checkmark & \checkmark & +2.82 & +0.00 & +0.00 & +0.00 & +0.70 \\
Single view & \checkmark & \checkmark &  & \checkmark & +2.43 & +0.30 & +1.68 & +0.67 & +1.27 \\
Conservative exec. & \checkmark & \checkmark & \checkmark &  & +3.73 & +2.47 & +4.05 & +7.06 & +4.33 \\
\rowcolor{SkyBlue!12}
\method{} & \checkmark & \checkmark & \checkmark & \checkmark & \textbf{+6.72} & \textbf{+4.19} & \textbf{+9.49} & \textbf{+9.67} & \textbf{+7.52} \\
\bottomrule
\end{tabular}}
\vspace{-2mm}
\end{table}

\noindent
\begin{minipage}[t]{0.48\linewidth}
\centering
\footnotesize
\setlength{\tabcolsep}{2.5pt}
\captionof{table}{\textbf{Evidence interface.} Free-form asks the VLM to name a label, full label list exposes the whole vocabulary, and closed top-$k$ contains the current/base label, up to $k$ non-base score-derived alternatives, and \texttt{uncertain}.}
\label{tab:evidence_interface_main}
\vspace{1mm}
\resizebox{\linewidth}{!}{
\begin{tabular}{lrrrrr}
\toprule
Interface & VOC21 & PC60 & City & ADE & Avg. \\
\midrule
Free-form & +2.82 & +0.00 & +0.00 & +0.00 & +0.70 \\
Full label list & +5.97 & +0.55 & +6.02 & +6.56 & +4.78 \\
Closed top-1 & +5.68 & +1.38 & +7.95 & +8.43 & +5.86 \\
Closed top-2 & \textbf{+6.94} & +2.22 & +6.79 & +8.93 & +6.22 \\
\rowcolor{SkyBlue!12}
Closed top-3 & +6.72 & \textbf{+4.19} & \textbf{+9.49} & \textbf{+9.67} & \textbf{+7.52} \\
\bottomrule
\end{tabular}}
\end{minipage}
\hfill
\begin{minipage}[t]{0.48\linewidth}
\centering
\footnotesize
\captionof{figure}{\textbf{Execution frontier.} Bars show $\Delta$ mIoU on VOC, PC, and ADE; the dashed line reports certified support. \method{} reaches the strongest gains by turning broader verified evidence into executable propagation.}
\label{fig:execution_policy_main}
\vspace{1mm}
\includegraphics[width=0.96\linewidth]{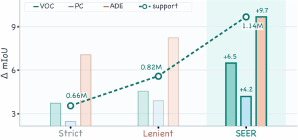}
\end{minipage}
\vspace{1mm}

\noindent
\begin{minipage}[t]{0.49\linewidth}
\centering
\footnotesize
\setlength{\tabcolsep}{3pt}
\captionof{table}{\textbf{Matched-input controls.} Every row uses the same frozen prediction, proposals, candidates, audit pool, and 2,000-audit cap.}
\label{tab:fair_nonvlm_main}
\vspace{1mm}
\resizebox{\linewidth}{!}{
\begin{tabular}{lrrrrr}
\toprule
Control & VOC21 & PC60 & City & ADE & Avg. \\
\midrule
OVSS score rule & +2.10 & +3.80 & +2.53 & +2.13 & +2.64 \\
CLIP crop classifier & +3.88 & +2.38 & +2.70 & +5.80 & +3.69 \\
VLM + score support & +2.09 & +0.87 & +1.74 & +2.11 & +1.70 \\
Dense-CRF smoothing & +0.01 & $-$0.01 & +0.02 & +0.01 & +0.01 \\
\rowcolor{SkyBlue!12}
\method{} & \textbf{+6.72} & \textbf{+4.19} & \textbf{+9.49} & \textbf{+9.67} & \textbf{+7.52} \\
\bottomrule
\end{tabular}}
\end{minipage}
\hfill
\begin{minipage}[t]{0.49\linewidth}
\centering
\footnotesize
\setlength{\tabcolsep}{3pt}
\captionof{table}{\textbf{Multi-view evidence.} Results use the default audit pool, verifier, candidates, gate, and propagation policy.}
\label{tab:three_view_main}
\vspace{1mm}
\resizebox{\linewidth}{!}{
\begin{tabular}{lrrrrr}
\toprule
View & VOC21 & PC60 & City & ADE & Avg. \\
\midrule
Full image & +4.26 & +2.18 & +4.43 & +4.71 & +3.90 \\
Local crop & +4.91 & +2.76 & +5.88 & +4.64 & +4.55 \\
Tight crop & +3.48 & +2.07 & +3.96 & +3.52 & +3.26 \\
\rowcolor{SkyBlue!12}
Three views & \textbf{+6.72} & \textbf{+4.19} & \textbf{+9.49} & \textbf{+9.67} & \textbf{+7.52} \\
\bottomrule
\end{tabular}}
\end{minipage}
\vspace{1mm}

\paragraph{Evidence interface and execution.}
The interface study in Table~\ref{tab:evidence_interface_main} shows why the verifier should operate in a bounded label space. Free-form answers may miss the active vocabulary, while the full vocabulary introduces legal but weakly supported choices. With the default top-3 alternatives, the analysis-only target-in-set coverage is 99.9\% across the four mechanism benchmarks. Filtering raises decision precision from 57.9\% before execution to 82.3\% among executed answers; 36.2\% of audits leave the prediction unchanged and only 0.2\% are invalid or out of set (Appendix~\ref{app:closed_set}). Figure~\ref{fig:execution_policy_main} gives the matching pixel-support story: strict routing is safe but low-recall, whereas SEER expands only high-confidence, score-compatible regions.

\paragraph{Controls and visual context.}
Under the matched inputs in Tab.~\ref{tab:fair_nonvlm_main}, a score rule, ordinary CLIP crop classifier, VLM answer with only score support, and Dense-CRF smoothing all remain below full SEER. The gain therefore is not reproduced by proposals plus a heuristic, a stronger regional classifier, an isolated VLM answer, or generic smoothing alone. The view ablation in Tab.~\ref{tab:three_view_main} further shows that a tight crop is insufficient: global scene context, local boundaries, and region appearance provide complementary evidence.

\noindent
\begin{minipage}[t]{0.57\linewidth}
\centering
\scriptsize
\setlength{\tabcolsep}{2.1pt}
\captionof{table}{\textbf{VLM auditor transfer.} All rows use the same replay protocol and differ only in the regional verifier.}
\label{tab:vlm_family_main}
\vspace{1mm}
\resizebox{\linewidth}{!}{
\begin{tabular}{llrrrrr}
\toprule
Family & Scale & VOC21 & PC60 & City & ADE & Avg. \\
\midrule
Qwen & 0.8B & +2.23 & +0.34 & +0.70 & +2.64 & +1.48 \\
Qwen & 4B & +5.86 & +3.26 & +7.49 & +8.45 & +6.26 \\
\rowcolor{SkyBlue!12}
Qwen & 9B & \textbf{+6.72} & \textbf{+4.19} & \textbf{+9.49} & \textbf{+9.67} & \textbf{+7.52} \\
LLaVA-OV-1.5 & 4B & +5.44 & +2.30 & +6.46 & +6.21 & +5.10 \\
LLaVA-OV-1.5 & 8B & +6.01 & +3.72 & +8.35 & +8.52 & +6.65 \\
InternVL3 & 1B & +1.38 & +1.41 & +1.43 & +1.25 & +1.37 \\
InternVL3 & 9B & +5.38 & +3.91 & +7.43 & +7.93 & +6.16 \\
Molmo2 & 4B & +5.17 & +2.65 & +6.28 & +6.63 & +5.18 \\
Molmo2 & 9B & +5.64 & +3.84 & +8.77 & +8.95 & +6.80 \\
\bottomrule
\end{tabular}}
\end{minipage}
\hfill
\begin{minipage}[t]{0.39\linewidth}
\centering
\captionof{figure}{\textbf{Audit budget.} More proposals improve repair quality; SEER stays above score-gated execution.}
\label{fig:topk_budget_main}
\vspace{1mm}
\includegraphics[width=\linewidth]{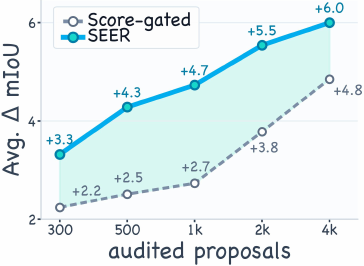}
\end{minipage}
\subsection{Cost and deployment scope}
\label{sec:cost_visual}
\begin{table}[H]
\centering
\footnotesize
\caption{\textbf{End-to-end cost (s/image).} Validation-weighted measurements on the same hardware. Cached SEER reuses fixed SAM2 proposals; cold SEER includes proposal generation. Audit load averages 0.71 calls per image.}
\label{tab:efficiency}
\vspace{-1mm}
\setlength{\tabcolsep}{3.6pt}
\begin{tabular}{lrrrrrr}
\toprule
Configuration & Proposal gen. & OVSS & Audit & Writeback & Total & Overhead \\
\midrule
Frozen OVSS & -- & 5.042 & -- & -- & 5.042 & -- \\
Cached SEER & -- & 5.042 & 0.424 & 0.043 & 5.509 & +9.3\% \\
Cold SEER & 0.508 & 5.042 & 0.424 & 0.043 & 6.018 & +19.3\% \\
\bottomrule
\end{tabular}
\end{table}

\vspace{1mm}
\noindent
\begin{minipage}[t]{0.62\linewidth}
\vspace{0pt}
\centering
\includegraphics[width=\linewidth]{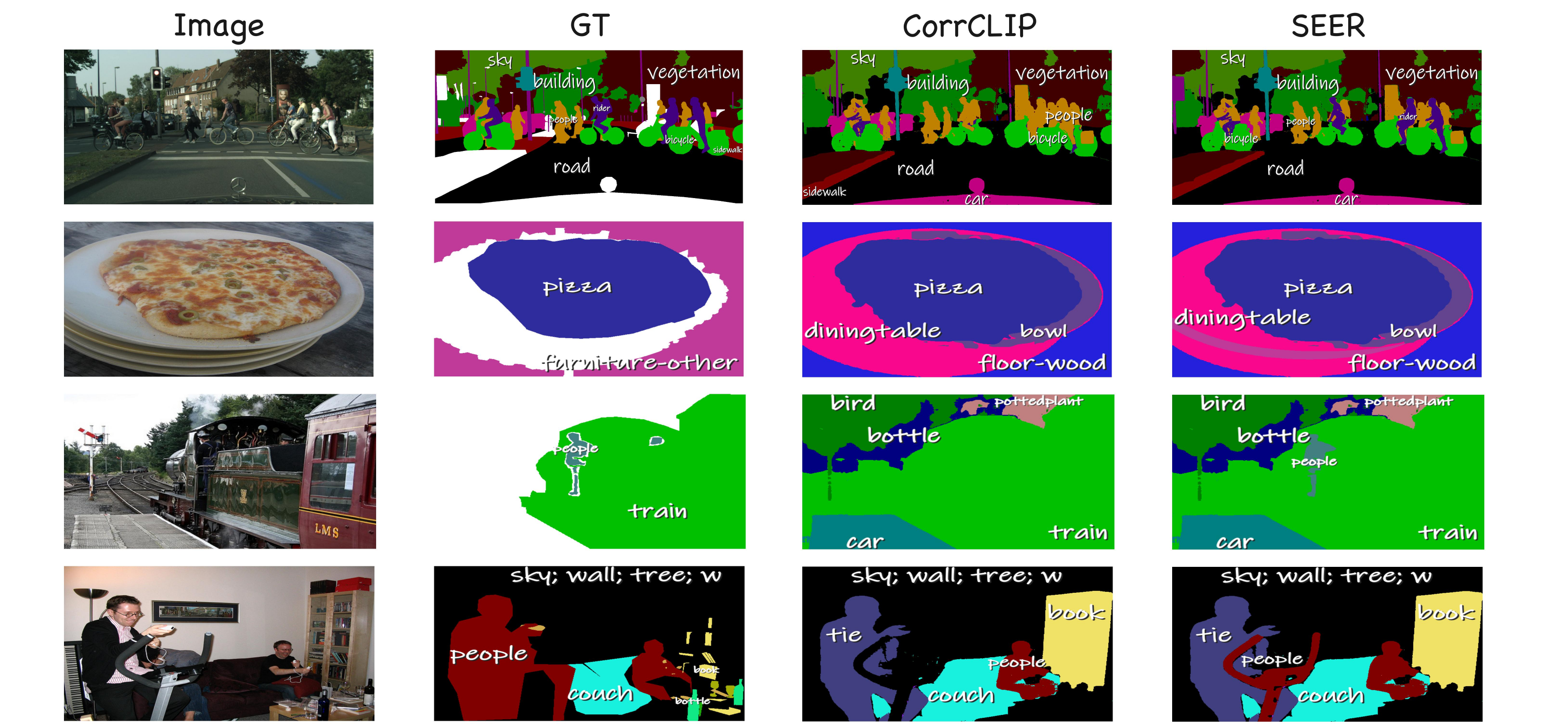}
\captionof{figure}{\textbf{Residual failure cases.} Mixed proposals, thin structures, and weak score support can leave errors near boundaries. Additional examples appear in Appendix~\ref{app:visuals}.}
\label{fig:failure_main}
\end{minipage}
\hfill
\begin{minipage}[t]{0.34\linewidth}
\vspace{0pt}
\small
\raggedright
\textbf{Robustness checks.}
Joint threshold scaling changes the gain by at most 0.14 mIoU. One-factor sweeps give a largest per-dataset range of 0.083 mIoU. Replaying 8,000 decisions with three shuffled candidate orders gives a largest per-dataset decline of 0.078 mIoU. These checks use cached verifier outputs and leave the remaining pipeline fixed.

\textbf{Failure boundary.}
Residual errors concentrate in mixed proposals, thin structures, candidate misses, and fragmented score support. The gate and conservative route reduce harmful writes, while some difficult regions remain unchanged.
\end{minipage}

\paragraph{Proposal and execution audit.}
We further audit 2,000 selected SAM2 proposals, with 500 from each mechanism benchmark. Ground truth is used only for this post-hoc analysis after proposal selection, VLM decisions, and execution have been fixed. Table~\ref{tab:proposal_execution_audit} separates proposal quality from the behavior of the executable gate.

\begin{table}[H]
\centering
\footnotesize
\caption{\textbf{Proposal quality and execution coverage.} Purity and semantic correctness are analysis-only measurements over the fixed audit pool.}
\label{tab:proposal_execution_audit}
\vspace{-1mm}
\setlength{\tabcolsep}{4pt}
\begin{tabular}{lrl}
\toprule
Diagnostic & Observation & Execution outcome \\
\midrule
GT purity $\geq 0.95$ & 1,732 / 2,000 (86.6\%) & +0.466 mean mIoU \\
GT purity $<0.70$ & 71 / 2,000 (3.55\%) & 43 / 71 filtered \\
Low-purity bin effect & $-0.0055$ / $+0.0018$ & Near-zero aggregate effect \\
Correct high-confidence decisions & 1,397 & 1,061 base confirmations \\
Correct non-base directions & 336 & 327 executed (97.32\%) \\
Empty supported execution & 0 & 9 safely retain $Y^0$ \\
\bottomrule
\end{tabular}
\end{table}

The useful gain is concentrated in high-purity proposals, while most low-purity proposals are suppressed and have negligible aggregate effect. Among correct non-base directions, the score-side executor applies 327 of 336 decisions; the remaining nine preserve the initial prediction, and none fails through an empty support. This analysis addresses proposal contamination and overly conservative execution separately.

\section{Conclusion and limitations}
\label{sec:conclusion}
\method{} frames OVSS repair as executable region propagation after a fixed prediction has been produced. It couples closed-set VLM evidence with OVSS score-space execution: the verifier identifies actionable regional labels, and the segmenter decides their pixel support. This evidence-to-execution design supports high-recall propagation on coherent regions while using conservative support near uncertain boundaries. Across eight benchmarks, cross-base plug-in tests, and protocol-matched controls, the experiments show consistent repair gains with controlled proposal-level intervention cost. The controls attribute the gain to the coupling of closed-set regional evidence and score-controlled execution; proposals, crop classification, and smoothing each remain below the full system. The complete system is intended for offline or selective repair. It adds 9.3\% cached or 19.3\% cold latency and uses 18.7GB peak verifier memory in the measured configuration. Proposal misses, absent candidates, verifier errors, and weak score support remain important limitations, especially for thin or mixed regions. Future work will study lighter auditing, stronger proposal coverage, and more robust boundary execution.

\clearpage

\subsection*{Reproducibility statement}
Appendix~\ref{app:implementation} documents the execution contract, prompts, decoding settings, preprocessing, dataset mappings, and evaluation protocol. It also reports matched-input controls, threshold and candidate-order sensitivity, verifier and proposal-source transfer, and complete resource measurements. All benchmarks and frozen models are public, and the same parameter profile is used throughout.

\bibliographystyle{iclr2027_conference}
\bibliography{references}

\clearpage
\maketitlesupplementary
\appendix

\section{Supplementary Overview}
\label{app:overview}
The main paper reports the compact experimental chain supporting the central claim that \method{} converts closed-set regional evidence into executable OVSS label propagation. The supplementary material gives the extended diagnostics behind that chain and uses the same terminology as the main text. \textit{Coherent region propagation} applies a certified label to the full proposal support. \textit{Boundary-conservative propagation} restricts the support to score- and margin-supported pixels. \textit{Adaptive propagation} chooses between these regimes based on evidence form and predicted risk.

\section{Theory and Algorithmic Summary}
\label{app:theory_algorithm}
This section gives the formal reading behind the main text. It does not claim a universal performance guarantee; instead, it shows why a VLM-assisted OVSS repair action naturally decomposes into a label-space direction and a pixel-space projector.

\paragraph{Notation.}
Let $\Omega$ be the pixel lattice, $n=|\Omega|$, and $\mathcal{C}=\{1,\ldots,C\}$ be the active OVSS label space. At repair step $t$, the prediction map is $Y^t\in\mathcal{C}^{n}$ and its one-hot matrix is $Z^t\in\{0,1\}^{n\times C}$, where the $p$-th row is $z_p^t=e_{Y_p^t}$. For analysis only, let $Y^\star$ be the ground-truth map, $g_p=Y^\star_p$, and $Z^\star$ its one-hot matrix. A local repair action is
\begin{equation}
    r=(\ell,W),\qquad \ell\in\mathcal{C},\quad W\subseteq\Omega,
\label{eq:supp_action_def}
\end{equation}
where $\ell$ is the target label and $W$ is the set of pixels to be changed. Let $\mathbf{1}\in\mathbb{R}^{n}$ be the all-one vector and let $P_W=\operatorname{diag}(\mathbf{1}_W)$ be the diagonal projector onto pixels in $W$. Applying $r$ is the linear update
\begin{equation}
Z^{t+1}=T_{\ell,W}(Z^t)
=Z^t+P_W(\mathbf{1}e_\ell^\top-Z^t).
\label{eq:supp_projected_update}
\end{equation}
Thus $e_\ell$ is a direction in label space, while $P_W$ is a projector in pixel space. This separation is the algebraic object that SEER operationalizes.

\begin{proposition}[Exact benefit--risk decomposition]
\label{prop:benefit_risk}
Define the pixel-error loss $L(Z)=n-\langle Z,Z^\star\rangle_F$. For any action $r=(\ell,W)$,
\begin{equation}
\Delta L(r)
:=L(T_{\ell,W}(Z^t))-L(Z^t)
=\mathcal{H}(\ell,W)-\mathcal{B}(\ell,W),
\label{eq:supp_delta_hb}
\end{equation}
where
\begin{equation}
\mathcal{H}(\ell,W)=\sum_{p\in W}\mathbf{1}[Y^t_p=g_p]\mathbf{1}[\ell\neq g_p],
\qquad
\mathcal{B}(\ell,W)=\sum_{p\in W}\mathbf{1}[Y^t_p\neq g_p]\mathbf{1}[\ell=g_p].
\label{eq:supp_hb_def}
\end{equation}
Here $\mathcal{H}$ counts pixels harmed by the action and $\mathcal{B}$ counts pixels repaired by the action.
\end{proposition}
\begin{proof}
Starting from the definition of $L$, the change in error is
\begin{align}
\Delta L(r)
&=\big(n-\langle Z^{t+1},Z^\star\rangle_F\big)-\big(n-\langle Z^t,Z^\star\rangle_F\big) \\
&=-\left\langle Z^{t+1}-Z^t,Z^\star\right\rangle_F \\
&=-\left\langle P_W(\mathbf{1}e_\ell^\top-Z^t),Z^\star\right\rangle_F \\
&=\left\langle P_W(Z^t-\mathbf{1}e_\ell^\top),Z^\star\right\rangle_F .
\label{eq:supp_delta_derivation}
\end{align}
Because $P_W$ is diagonal, only rows in $W$ contribute:
\begin{align}
\Delta L(r)
&=\sum_{p\in W}\left\langle z_p^t-e_\ell,e_{g_p}\right\rangle \\
&=\sum_{p\in W}\big(\mathbf{1}[Y_p^t=g_p]-\mathbf{1}[\ell=g_p]\big).
\label{eq:supp_pixel_sum}
\end{align}
For a selected pixel $p$, the summand is $+1$ exactly when $Y_p^t=g_p$ and $\ell\neq g_p$, i.e., a correct pixel is overwritten by a wrong label; it is $-1$ exactly when $Y_p^t\neq g_p$ and $\ell=g_p$, i.e., an incorrect pixel is repaired; otherwise it is $0$. Grouping the positive and negative terms gives Eqs.~\ref{eq:supp_delta_hb}--\ref{eq:supp_hb_def}.
\end{proof}

Proposition~\ref{prop:benefit_risk} explains why regional recognition alone cannot determine the value of a repair. Even if the label direction $e_\ell$ is semantically correct for part of a proposal, the action can still increase error when the projector $P_W$ covers already-correct pixels. The central quantity is the net benefit produced by the pair $(e_\ell,P_W)$.

\begin{lemma}[Regional evidence alone cannot identify the support projector]
\label{lem:non_identifiability}
Let an evidence-only executor be any deterministic rule $G$ that receives only a proposal $M$ and a regional evidence tuple $a$, and returns a support $W=G(M,a)\subseteq M$. Suppose $a$ certifies the same target label $\ell$ for two local configurations but does not include dense OVSS scores, margins, or pixel ownership information. Then the support is not identifiable from $(M,a)$: there exist two configurations consistent with the same evidence for which no single $W$ minimizes $\Delta L$ in both configurations.
\end{lemma}
\begin{proof}
Assume, for contradiction, that the support projector is identifiable from $(M,a)$, i.e., there exists a rule $G$ whose support is optimal for every configuration consistent with the same regional evidence. Choose a boundary pixel $p\in M$ whose regional evidence supports label $\ell$, and keep all pixels except $p$ identical across two configurations. Construct configuration $A$ so that $p$ is already correct and should not be edited:
\begin{equation}
Y^{t,A}_p=g^A_p\neq \ell .
\label{eq:supp_config_a}
\end{equation}
Construct configuration $B$ so that the same pixel is a residual error and should be changed to $\ell$:
\begin{equation}
Y^{t,B}_p\neq g^B_p=\ell .
\label{eq:supp_config_b}
\end{equation}
Both configurations share the same proposal $M$ and the same regional evidence tuple $a$, because $a$ only states that the region provides evidence for label $\ell$; it does not encode whether the boundary pixel $p$ is already correct or belongs to the repairable part. Therefore $G$ must make the same inclusion decision for $p$ in both configurations.

If $p\in G(M,a)$, then by Eq.~\ref{eq:supp_hb_def} this pixel contributes $+1$ to $\Delta L$ in configuration $A$, while excluding it would contribute $0$ and strictly improve the action. Hence $G$ is not optimal in $A$. If $p\notin G(M,a)$, then this pixel contributes $0$ in configuration $B$, while including it would contribute $-1$ and strictly improve the action. Hence $G$ is not optimal in $B$. Both cases contradict uniform optimality. Thus regional evidence alone cannot identify the support projector.
\end{proof}

\begin{corollary}[Evidence-to-execution factorization]
\label{cor:evidence_execution}
A repair interface that aims to reduce $\Delta L$ should use different information for the two factors of an action:
\begin{equation}
\underbrace{e_\ell}_{\text{label direction}}
\leftarrow f_{\mathrm{vlm}}(I,M_i,\mathcal{A}_i),
\qquad
\underbrace{P_{W_i}}_{\text{support projector}}
\leftarrow \Gamma(S,Y^0,M_i,\ell,a_i).
\label{eq:supp_factorized_action}
\end{equation}
This is the formal motivation for SEER. The VLM verifies the semantic direction, while the OVSS score field and propagation policy determine the executable projector.
\end{corollary}

\paragraph{Closed candidates as a label-space projection.}
For proposal $M_i$, let $b_i$ be the dominant current label and let $\mathcal{Q}_i$ be the score-derived alternatives. The token \texttt{uncertain} represents abstention and is excluded from the class set. We define the executable candidate set
\begin{equation}
\mathcal{A}_i^{+}=\{b_i\}\cup\mathcal{Q}_i\subseteq\mathcal{C}.
\label{eq:supp_exec_candidates}
\end{equation}
It induces a coordinate subspace and projection operator
\begin{equation}
\mathcal{U}_i=\operatorname{span}\{e_c:c\in\mathcal{A}_i^{+}\}\subseteq\mathbb{R}^{C},
\qquad
\Pi_{\mathcal{U}_i}v=\sum_{c\in\mathcal{A}_i^{+}}\langle v,e_c\rangle e_c .
\label{eq:supp_candidate_projection}
\end{equation}
If $v_i\in\mathbb{R}^{C}$ denotes the verifier's latent regional compatibility over the active vocabulary, a full-vocabulary verifier searches $\arg\max_{c\in\mathcal{C}} v_{i,c}$, while SEER searches only within the projected subspace:
\begin{equation}
\hat{\ell}_i
\in\arg\max_{c\in\mathcal{A}_i^{+}} \langle \Pi_{\mathcal{U}_i}v_i,e_c\rangle,
\quad \text{or abstains with }\texttt{uncertain}.
\label{eq:supp_closed_argmax}
\end{equation}
This makes the closed candidate interface more than prompt shortening. It reduces the executable action directions from $C$ to $|\mathcal{A}_i^{+}|$, and every non-abstained direction is both legal in the active label space and supported by the OVSS score tensor. The evidence-interface ablations in Tables~\ref{tab:evidence_interface_main} and~\ref{tab:supp_closed_set_abs} empirically support this projection view.

\paragraph{Propagation as constrained projector selection.}
For a certified label $\ell$, define the score-feasible support family
\begin{equation}
\mathfrak{P}_i(\ell;S,Y^0)
=\{P_W:
W\subseteq M_i,
\rho(p,\ell,b_i)=1\ \text{for all }p\in W\},
\label{eq:supp_projector_family}
\end{equation}
where $\rho$ is the score-support predicate in Eq.~\ref{eq:score_support}. If $Z^\star$ were known, the best support within this family would solve
\begin{equation}
P_i^\star(\ell)
\in
\arg\min_{P_W\in\mathfrak{P}_i(\ell;S,Y^0)}
\left\langle P_W(Z^t-\mathbf{1}e_\ell^\top),Z^\star\right\rangle_F .
\label{eq:supp_oracle_projector}
\end{equation}
Since $Z^\star$ is unavailable at inference time, SEER uses $S$, margins, evidence form, and risk predicates as an executable surrogate for selecting $P_W$. Coherent propagation chooses a larger projector after strong certification, while score-grounded and boundary-conservative propagation choose smaller projectors when score support is ambiguous. The resulting action is therefore a constrained selector in the OVSS score space.

\begin{algorithm}[H]
\caption{\textbf{SEER inference.} Regional evidence is certified before pixel-level propagation.}
\label{alg:seer}
\small
\begin{algorithmic}[1]
\Require image $I$, fixed prediction $Y^0$, score tensor $S$, label space $\mathcal{C}$, SAM2 proposal bank $\mathcal{M}$, audit budget $K$
\Ensure repaired prediction $Y$
\State $Y \gets Y^0$
\State $\mathcal{R}(I) \gets \operatorname{TopK}_{M_i\in\mathcal{M}}\pi(Y^0,S,M_i)$
\For{each $M_i\in\mathcal{R}(I)$}
    \State $b_i \gets \operatorname{mode}_{p\in M_i}Y_p$
    \State $\mathcal{A}_i \gets \{b_i\}\cup\mathcal{Q}_i(S,Y,M_i)\cup\{\texttt{uncertain}\}$
    \State $a_i=(\hat{\ell}_i,\mathrm{conf}_i,\mathrm{form}_i)\gets f_{\mathrm{vlm}}(I,M_i,\mathcal{A}_i)$
    \If{$a_i$ is not executable under Eq.~\ref{eq:execution_gate}}
        \State \textbf{continue}
    \EndIf
    \State $\ell_i\gets\hat{\ell}_i$
    \State $W_i\gets \Gamma(S,Y,M_i,\ell_i,a_i)$
    \State $Y_p\gets \ell_i$ for all $p\in W_i$
\EndFor
\State \Return $Y$
\end{algorithmic}
\end{algorithm}

\section{Implementation Details}
\label{app:implementation}
\paragraph{Unified protocol.}
The protocol matrix in Table~\ref{tab:unified_protocol} records which factor changes in each study. The default state uses a fixed SAM2 proposal bank, closed top-3 candidates with \texttt{uncertain}, local Qwen3.5-9B deterministic decoding, the three-view prompt, and the default SEER propagation policy. The same coefficients and thresholds are used for all benchmarks without dataset-specific tuning. Closed top-$k$ denotes the current/base label plus up to $k$ non-base score-derived alternatives; \texttt{uncertain} is an abstention output and is not counted as a semantic candidate.

\begin{table}[H]
\centering
\footnotesize
\setlength{\tabcolsep}{3.6pt}
\renewcommand{\arraystretch}{1.06}
\caption{\textbf{Unified protocol.} Each row lists the experimental scope, the factor varied, and the quantities kept identical.}
\label{tab:unified_protocol}
\vspace{1mm}
\begin{tabular}{@{}>{\raggedright\arraybackslash}p{0.12\linewidth}
                >{\raggedright\arraybackslash}p{0.19\linewidth}
                >{\raggedright\arraybackslash}p{0.19\linewidth}
                >{\raggedright\arraybackslash}p{0.40\linewidth}@{}}
\toprule
Study & Scope & Varied factor & Held fixed \\
\midrule
Main & Full validation & Backbone; benchmark & Default verifier, prompt, candidates, policy \\
Components & Four benchmarks & Module ablation & Prediction, proposals, candidates, parser, split \\
Interface & Default audit pool & Response format & Prediction, proposals, executable gate, policy \\
Views & Default audit pool & Visual context & Candidates, verifier, executable gate, policy \\
Cross-base & Full pipeline & OVSS base & Same coefficients, thresholds, verifier, and rules \\
Budget & Ranked prefixes & Audit budget & Cached evidence, executable gate, policy \\
\bottomrule
\end{tabular}
\vspace{-1mm}
\end{table}

\begin{table}[H]
\centering
\small
\setlength{\tabcolsep}{4.5pt}
\renewcommand{\arraystretch}{1.12}
\caption{\textbf{Deterministic execution contract.} Fixed constants and routing rules are shared by all benchmarks; invalid or abstained audits never modify pixels.}
\label{tab:exact_constants_rules}
\vspace{1mm}
\begin{tabular}{@{}>{\raggedright\arraybackslash}p{0.17\linewidth}
                >{\raggedright\arraybackslash}p{0.34\linewidth}
                >{\raggedright\arraybackslash}p{0.43\linewidth}@{}}
\toprule
Component & Fixed rule/value & Role in execution \\
\midrule
Priority score & $\pi_i=1.20\log(1+100\alpha_i)+q_i+0.75\max(\delta_i,-0.35)+0.45\lambda_i+0.20\eta_i-0.90r_i$ & Ranks proposals before sparse VLM audit; weights are fixed once and not benchmark-fitted. \\
Risk predicate & Object-like candidate/base, $r_i>0.20$, or $\alpha_i<0.015$ & Blocks broad writes for risky, small, or boundary-sensitive proposals. \\
Score support & $(\tau_s,\tau_m,\tau_l)=(0.02,0.20,0.25)$ in Eq.~\ref{eq:score_support} & Chooses score-compatible pixels for conservative propagation. \\
Propagation route & Broad support for coherent low-risk evidence; conservative support for object, boundary, small, or risky evidence & Keeps boundary evidence executable but restricts writes to selected support. \\
Parser and gate & Invalid JSON, out-of-set label, \texttt{uncertain}, base label, low/medium confidence, \texttt{mixed}, or \texttt{unclear} & Abstains with no pixel update; deterministic parser, no retry. \\
VLM audit & Local Qwen3.5-9B; three-view prompt; constrained JSON; deterministic decoding; 96-token cap & One audited proposal gives one sparse VLM call; no closed oracle model is used. \\
View builder & Full image width 640; local crop fits $448^2$; tight crop fits $336^2$ & Uses the same global, neighborhood, and tight views for every audit. \\
SAM2 proposals & Hiera-L fp16; points 8; IoU/stability 0.4; multimask off; crop layers 0; NMS 0.7 & Supplies spatial domains only, never semantic predictions. \\
Audit budget & 2,000 unique audits per benchmark & $K$ counts VLM calls, not images or labels. \\
\bottomrule
\end{tabular}
\vspace{-3mm}
\end{table}

\begin{table}[H]
\centering
\small
\setlength{\tabcolsep}{3pt}
\caption{\textbf{Proposal-source diagnostic.} The main protocol uses SAM2 to align with the CorrCLIP-style configuration; this diagnostic varies only the mask generator that supplies spatial domains.}
\label{tab:proposal_source_scope}
\vspace{1mm}
\resizebox{\linewidth}{!}{
\begin{tabular}{lccccccccc}
\toprule
Mask generator & VOC21 & VOC20 & PC59 & PC60 & City & ADE & Stuff & Object & Avg. \\
\midrule
Mask2Former~\cite{mask2former} & 78.5 & 93.5 & 54.8 & 47.6 & 53.9 & 36.7 & 40.1 & 54.6 & 57.5 \\
\rowcolor{SkyBlue!12}
SAM2~\cite{sam2} & 79.6 & 94.2 & 55.2 & 48.2 & 56.9 & 38.3 & 38.6 & 52.9 & 58.0 \\
EntitySeg~\cite{entityseg} & 81.1 & 94.8 & 57.8 & 49.7 & \textbf{58.1} & \textbf{40.6} & 39.6 & 54.2 & 59.5 \\
SAM3~\cite{sam3} & \textbf{82.3} & \textbf{95.6} & \textbf{58.6} & \textbf{50.5} & 57.9 & 40.0 & \textbf{40.4} & \textbf{56.5} & \textbf{60.2} \\
\bottomrule
\end{tabular}}
\vspace{-1mm}
\end{table}

\method{} uses prediction-side quantities at inference time. The SAM2 proposal bank is precomputed and fixed in the main protocol to match the CorrCLIP-style evaluation setup; Table~\ref{tab:proposal_source_scope} varies only this spatial-domain source. Proposals define candidate supports for auditing and propagation, while semantic labels come from the OVSS score tensor and closed-set regional evidence. Fixing SAM2 keeps the protocol aligned and isolates the repair layer itself.

This diagnostic supports two points. First, Mask2Former, EntitySeg, and SAM3 all keep the pipeline functional when they provide reasonable spatial support. Second, stronger proposal coverage can lift the ceiling because proposals define the largest regions that can be audited and certified. The main experiments fix SAM2 for protocol alignment. SAM3 is strongest here, while its prompt-conditioned masks make it a diagnostic alternative outside the default protocol.

In the priority score, $\alpha_i=|M_i|/|\Omega|$, $q_i$ is the dominant-label fraction, $\delta_i$ is the candidate-minus-base score gap, $\lambda_i$ is the low-margin fraction, $\eta_i$ is the mean entropy signal when available, and $r_i$ is the fixed risk-label fraction. For the view builder, $d=\max(\mathrm{box\ width},\mathrm{box\ height})$. The full view resizes the highlighted image to width 640, the local crop expands the proposal box by $\max(32\mathrm{px},0.3d)$ before fitting it inside $448\times448$, and the tight crop expands by $\max(6\mathrm{px},0.3d)$ before fitting it inside $336\times336$. The object-like routing set includes people/person, vehicles, animals, common furniture, and small handled objects when those names appear in the active dataset vocabulary. The structural risk-ratio set is fixed to common boundary-sensitive labels including background, wall, ceiling, floor, ground, road, building, tree, grass, sky, water, mountain, sidewalk, platform, rock, snow, wood, fence, door, window, curtain, light, sign, track, and shelves. These lists only affect routing and priority; they do not add candidate labels.

For a proposal $M_i$, the priority score combines four monotone signals: (i) the purity of the dominant baseline prediction within the proposal, (ii) proposal area normalized by image area, (iii) competition between the strongest non-baseline score and the baseline score, and (iv) the fraction of pixels predicted as risk-sensitive classes. This priority serves as a budget controller. Its coarse weights encode a fixed preference for impact, purity, score competition, uncertainty, and risk avoidance; they are not fitted per benchmark. It only orders proposals before VLM verification, and Appendix~\ref{app:priority} shows that ranking alone is insufficient without closed-set evidence and executable propagation.

For score-grounded propagation, Eq.~\ref{eq:score_support} uses three fixed constants. $\tau_s$ is a small tolerance that allows the target score to match the current baseline score, $\tau_m$ defines low-margin pixels, and $\tau_l$ is the wider tolerance allowed only inside low-margin support. We set $(\tau_s,\tau_m,\tau_l)=(0.02,0.20,0.25)$ once and keep the values unchanged across benchmarks. These rules form a deterministic execution contract: semantic evidence proposes a legal direction, while score support and risk predicates decide where the direction may write. Boundary-conservative propagation additionally requires low margin and is triggered when score support is weak after semantic certification. The default adaptive policy selects coherent propagation for certified low-risk regions and falls back to score-grounded or boundary-conservative propagation for object-like, boundary-like, large-risk, or high-risk proposals.

\noindent
\begin{minipage}[t]{0.51\linewidth}
\vspace{0pt}
\centering
\includegraphics[width=\linewidth]{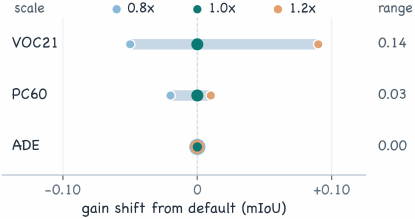}
\captionof{figure}{\textbf{Support-threshold stability.} Joint scaling around the default constants $(0.02,0.20,0.25)$ changes gains by at most 0.14 mIoU.}
\label{fig:tau_perturbation}
\end{minipage}
\hfill
\begin{minipage}[t]{0.45\linewidth}
\vspace{0pt}
\raggedright
This support-threshold replay varies fixed constants around their default values. Using the main closed top-3 reference, gains are $+6.72$, $+4.19$, and $+9.67$ on VOC21, PC60, and ADE. Joint scaling shifts the replayed gains by at most 0.14 mIoU; PC60 and ADE move by 0.03 and 0.00. One-factor sweeps over $\tau_s$, $\tau_m$, and $\tau_l$ change the four-dataset mean after-mIoU by 0.00007, 0.02093, and 0.01554, respectively; the largest range on any individual dataset is 0.08291. The audit interface, cached VLM evidence, parser, and propagation policy stay fixed in every replay.
\end{minipage}
\vspace{1mm}

\paragraph{VLM decoding.}
All VLM verifier outputs used by the main mechanism studies are produced with deterministic decoding (\texttt{do\_sample=False}; greedy decoding with zero or implicit temperature). The replay diagnostics therefore reuse cached verifier outputs exactly as generated, which removes sampling noise from the protocol and keeps the audit sequence reproducible.
The threshold stability diagnostic in Fig.~\ref{fig:tau_perturbation} checks the fixed support thresholds around their default setting with joint and one-factor perturbations.

\section{Closed-Set Evidence Diagnostics}
\label{app:closed_set}
The main paper reports evidence-interface gains using Qwen3.5-9B as the default verifier. The supplementary view adds absolute mIoU in Table~\ref{tab:supp_closed_set_abs}, execution counts in Table~\ref{tab:supp_closed_set_stats}, and VLM filtering behavior in Table~\ref{tab:vlm_execution_filtering}. These diagnostics should be read together because a stronger interface is useful only when it remains executable under the OVSS score field. Boundary-sensitive COCO-Object is analyzed separately in the routing and propagation diagnostics because its stuff-to-object label collapse makes broad regional evidence less directly executable, so the same label can be semantically plausible yet spatially risky.

\noindent
\begin{minipage}[t]{0.48\linewidth}
\centering
\footnotesize
\setlength{\tabcolsep}{2.4pt}
\captionof{table}{\textbf{Evidence interfaces.} Absolute mIoU for the main gain table.}
\label{tab:supp_closed_set_abs}
\vspace{1mm}
\resizebox{\linewidth}{!}{
\begin{tabular}{lrrrrr}
\toprule
Interface & VOC21 & PC60 & City & ADE & Avg. \\
\midrule
Initial prediction & 72.88 & 44.01 & 47.41 & 28.63 & 48.23 \\
Free-form label & 75.70 & 44.01 & 47.41 & 28.63 & 48.94 \\
Full label list & 78.85 & 44.56 & 53.43 & 35.19 & 53.01 \\
Closed top-1 & 78.56 & 45.39 & 55.36 & 37.06 & 54.09 \\
Closed top-2 & \textbf{79.82} & 46.23 & 54.20 & 37.56 & 54.45 \\
\rowcolor{SkyBlue!12}
Closed top-3 & 79.60 & \textbf{48.20} & \textbf{56.90} & \textbf{38.30} & \textbf{55.75} \\
\bottomrule
\end{tabular}}
\end{minipage}
\hfill
\begin{minipage}[t]{0.48\linewidth}
\centering
\footnotesize
\setlength{\tabcolsep}{2.7pt}
\captionof{table}{\textbf{Execution statistics.} Closed evidence executes useful regions.}
\label{tab:supp_closed_set_stats}
\vspace{1mm}
\resizebox{\linewidth}{!}{
\begin{tabular}{lrrrr}
\toprule
Interface & Gain & Exec. & Ben. px & Harm. px \\
\midrule
Free-form label & +0.70 & 455 & 3.49M & 0.04M \\
Full label list & +4.78 & 4,193 & 160.53M & 9.06M \\
Closed top-1 & +5.86 & 3,700 & 174.80M & 3.56M \\
Closed top-2 & +6.22 & 3,951 & 183.11M & 4.07M \\
\rowcolor{SkyBlue!12}
Closed top-3 & \textbf{+7.52} & \textbf{5,217} & \textbf{220.68M} & 5.11M \\
\bottomrule
\end{tabular}}
\end{minipage}
\vspace{2mm}

The candidate-size diagnostic in Table~\ref{tab:supp_candidate_behavior} compares the default Base + 3 interface with smaller and larger candidate lists. Target-in-set coverage is an analysis-only quantity computed with validation labels; it is unavailable to the verifier and propagation policy at inference time. All outcome rates use audited proposals as the denominator. Adding candidates changes both the verifier routing and abstention pattern, so correct outputs need not increase monotonically with $k$. The best gain comes from the balance between target coverage, repair outputs, execution filtering, and conservative exits. Top-3 increases repair and correct outputs while allowing more abstain/filter exits than Top-2, which is intended under richer candidate context.

\begin{table}[H]
\centering
\small
\setlength{\tabcolsep}{3.5pt}
\caption{\textbf{Candidate-set size and verifier outcomes.} Rates are averaged over the four mechanism benchmarks with 2,000 audited proposals each. Base + $k$ alt. denotes the current/base label plus up to $k$ non-base score-derived alternatives; \texttt{uncertain} is an abstention output. Target-in-set coverage is diagnosis-only. Repair and correct-output rates are normalized by all audited proposals; executed precision is reported separately in Table~\ref{tab:vlm_execution_filtering}.}
\label{tab:supp_candidate_behavior}
\vspace{1mm}
\resizebox{\linewidth}{!}{
\begin{tabular}{lrrrrrrr}
\toprule
Interface & Target-in-set cov. & Sem. cand. & Repair output & Correct output & Base/no edit & Abstain/filter & Avg. gain \\
\midrule
Base + 1 alt. & 56.3\% & 2.00 & 41.3\% & 40.7\% & 50.1\% & 8.6\% & +5.86 \\
Base + 2 alt. & 85.2\% & 2.97 & 47.6\% & 39.4\% & 46.3\% & 6.1\% & +6.22 \\
\rowcolor{SkyBlue!12}
Base + 3 alt. & \textbf{99.9\%} & \textbf{3.35} & \textbf{58.8\%} & \textbf{46.4\%} & \textbf{28.7\%} & 12.5\% & \textbf{+7.52} \\
\bottomrule
\end{tabular}}
\vspace{-1mm}
\end{table}

\begin{table}[H]
\centering
\small
\setlength{\tabcolsep}{4.5pt}
\caption{\textbf{Execution filtering of VLM evidence.} Candidate, filtered, and invalid/OOS rates use audited proposals as the denominator; raw accuracy and executed precision are measured on their corresponding pre- and post-filter subsets.}
\label{tab:vlm_execution_filtering}
\vspace{1mm}
\resizebox{\linewidth}{!}{
\begin{tabular}{lrrrrr}
\toprule
Benchmark & Cand. present & Raw high-conf. acc. & Executed precision & Filtered/no exec. & Invalid/OOS \\
\midrule
VOC21 & 100.0\% & 62.4\% & 81.9\% & 38.9\% & 0.2\% \\
PC60 & 99.5\% & 67.7\% & 92.7\% & 35.2\% & 0.4\% \\
City & 100.0\% & 48.5\% & 74.3\% & 36.2\% & 0.1\% \\
ADE & 100.0\% & 53.1\% & 80.2\% & 34.5\% & 0.1\% \\
\midrule
Average & 99.9\% & 57.9\% & 82.3\% & 36.2\% & 0.2\% \\
\bottomrule
\end{tabular}}
\vspace{-1mm}
\end{table}

These diagnostics support treating the VLM as a regional evidence source that still requires execution filtering. High-confidence answers alone have limited calibration. The parser, candidate-set validity checks, abstention path, and execution gate substantially improve the precision of decisions that are allowed to affect pixels. Table~\ref{tab:vlm_execution_filtering} measures audited proposals before and after filtering, while Table~\ref{tab:supp_failure_taxonomy} summarizes routed proposals; the denominators are therefore different.

\section{Propagation Policy Diagnostics}
\label{app:propagation}
The propagation replay is extended to all eight benchmarks in Table~\ref{tab:supp_propagation_full} and to pixel-level exposure in Table~\ref{tab:supp_propagation_pixels}. These endpoint policies show why propagation choice matters, not that one support type is universally superior. Coherent region propagation is strong when verified regions correspond to semantic units; boundary-conservative propagation is useful when object boundaries or proposal noise make broad propagation risky. The adaptive SEER policy is evaluated separately in Table~\ref{tab:supp_policy}, where routing among strict, lenient, and default support gives the strongest overall trade-off.

\begin{table}[H]
\centering
\small
\setlength{\tabcolsep}{3.5pt}
\caption{\textbf{Full propagation-policy replay over eight benchmarks.} Gains are measured over the fixed prediction. Beneficial/harmful pixels are diagnostic statistics computed after evaluation.}
\label{tab:supp_propagation_full}
\vspace{1mm}
\resizebox{\linewidth}{!}{
\begin{tabular}{lrrrrrrrrr}
\toprule
 & VOC20 & VOC21 & PC59 & PC60 & City & ADE & Stuff & Object & Avg. \\
\midrule
Initial prediction & 87.62 & 72.88 & 48.68 & 44.01 & 47.41 & 28.63 & 33.38 & 46.19 & 51.10 \\
Coherent mIoU & 91.92 & 77.38 & 53.18 & 46.91 & 53.61 & 35.03 & 36.38 & 49.19 & 55.45 \\
\rowcolor{SkyBlue!12}
Coherent gain & +4.30 & +4.50 & +4.50 & +2.90 & +6.20 & +6.40 & +3.00 & +3.00 & +4.35 \\
Boundary mIoU & 90.72 & 76.28 & 51.98 & 46.31 & 51.91 & 33.83 & 35.68 & 48.59 & 54.41 \\
Boundary gain & +3.10 & +3.40 & +3.30 & +2.30 & +4.50 & +5.20 & +2.30 & +2.40 & +3.31 \\
\bottomrule
\end{tabular}}
\end{table}

\begin{table}[H]
\centering
\small
\setlength{\tabcolsep}{3.5pt}
\caption{\textbf{Pixel exposure of propagation policies.} Harmful pixels are shown in millions; full mIoU gains, including COCO-Object, are reported in Table~\ref{tab:supp_propagation_full}.}
\label{tab:supp_propagation_pixels}
\vspace{1mm}
\resizebox{\linewidth}{!}{
\begin{tabular}{lrrrrrrr}
\toprule
 & VOC20 & VOC21 & PC59 & PC60 & City & ADE & Stuff \\
\midrule
Coherent harmful px & 0.02 & 1.32 & 1.14 & 13.85 & 1.41 & 0.76 & 1.18 \\
Boundary harmful px & 0.00 & 1.28 & 0.85 & 13.29 & 0.67 & 0.47 & 0.58 \\
Coherent beneficial px & 2.32 & 8.22 & 50.97 & 99.13 & 90.54 & 48.00 & 125.65 \\
Boundary beneficial px & 1.24 & 6.48 & 35.88 & 71.09 & 38.63 & 34.13 & 84.63 \\
\bottomrule
\end{tabular}}
\end{table}

\section{Region Prioritization Diagnostics}
\label{app:priority}
The priority module controls the audit budget. Its coefficients are coarse, fixed ordering preferences for area, purity, score competition, uncertainty, and risk exposure; they are not tuned per benchmark and are not used after a region has been audited. The replay in Table~\ref{tab:supp_priority_replay} reorders the same audited pool and reuses cached VLM decisions, isolating proposal ordering without new VLM calls. Area is useful, while the full priority is stronger on average. This supports a modest claim: prediction-side ranking exposes a compact auditable set, and executable evidence propagation produces the main gain.

\begin{table}[H]
\centering
\small
\setlength{\tabcolsep}{3pt}
\caption{\textbf{Audited-pool priority replay.} Values are mIoU gains over the fixed prediction when the same audited decisions are replayed under different proposal orderings.}
\label{tab:supp_priority_replay}
\resizebox{0.98\linewidth}{!}{
\begin{tabular}{lrrrrr}
\toprule
Benchmark & Random & Area-only & Uncertainty-only & Competition-only & Full priority \\
\midrule
VOC21 & +0.53 & +4.55 & +1.79 & +3.70 & \textbf{+6.72} \\
PC60 & +0.27 & +2.12 & +2.00 & +3.39 & \textbf{+4.19} \\
City & +1.32 & +4.73 & +1.64 & +3.55 & \textbf{+9.49} \\
ADE & +1.17 & +4.90 & +3.42 & +4.75 & \textbf{+9.67} \\
\midrule
Average & +0.82 & +4.08 & +2.21 & +3.85 & \textbf{+7.52} \\
\bottomrule
\end{tabular}}
\end{table}

\paragraph{Candidate-order stability.}
We replayed 8,000 cached decisions with the original candidate order and three independent shuffles. The corresponding after-mIoU values are 49.4254, 49.4820, 49.4558, and 49.4701. The largest decline on any individual dataset is 0.0778 mIoU. Candidate position therefore has little influence under deterministic decoding and the closed executable gate.

\section{Executable Gate Diagnostics}
\label{app:gate_efficiency}
The relaxed-gate diagnostic in Table~\ref{tab:supp_gate} removes the target/legal certification constraint and allows broader executable actions. Relaxing the gate can increase recall, while also expanding the number of writeback actions substantially and making conservative containment harder to preserve. The accompanying increase in routed support supports the gate's role as an executable certification constraint.

\begin{table}[H]
\centering
\small
\setlength{\tabcolsep}{3pt}
\caption{\textbf{Relaxing executable certification.} Removing the target/legal gate increases recall but permits many more writeback actions.}
\label{tab:supp_gate}
\begin{tabular}{lrrrr}
\toprule
Benchmark & Initial prediction & After relaxed gate & Gain & Executed actions \\
\midrule
VOC21 & 72.88 & 78.58 & +5.70 & 5,024 \\
PC60 & 44.01 & 49.59 & +5.58 & 7,572 \\
City & 47.41 & 56.56 & +9.15 & 1,955 \\
ADE & 28.63 & 38.21 & +9.58 & 1,975 \\
COCO-Object & 46.19 & 50.55 & +4.36 & 1,608 \\
\bottomrule
\end{tabular}
\end{table}

\section{Control Interfaces and Non-VLM Diagnostics}
\label{app:controls}
Protocol-matched controls are compared in Table~\ref{tab:supp_controls}, with pixel statistics in Table~\ref{tab:supp_controls_stats}. The score tensor contains useful repair signal, but score-only rules do not produce regional semantic evidence. A CLIP crop classifier tests ordinary region recognition, score-grounded support removes SEER's adaptive executable propagation, and Dense-CRF-style smoothing tests generic local regularization. All controls remain below SEER under the same fixed-prediction protocol.

\noindent
\begin{minipage}[t]{0.50\linewidth}
\centering
\footnotesize
\setlength{\tabcolsep}{2.4pt}
\captionof{table}{\textbf{Protocol-matched controls.} Same prediction state; only the repair signal or execution rule changes.}
\label{tab:supp_controls}
\vspace{1mm}
\resizebox{\linewidth}{!}{%
\begin{tabular}{@{}lrrrrr@{}}
\toprule
Method & VOC21 & PC60 & City & ADE & Avg. \\
\midrule
OVSS score rule & +2.10 & +3.80 & +2.53 & +2.13 & +2.64 \\
CLIP crop classifier & +3.88 & +2.38 & +2.70 & +5.80 & +3.69 \\
VLM score support & +2.09 & +0.87 & +1.74 & +2.11 & +1.70 \\
Dense-CRF smoothing & +0.01 & -0.01 & +0.02 & +0.01 & +0.01 \\
\rowcolor{SkyBlue!12}
\method{} & \textbf{+6.72} & \textbf{+4.19} & \textbf{+9.49} & \textbf{+9.67} & \textbf{+7.52} \\
\bottomrule
\end{tabular}}
\end{minipage}
\hfill
\begin{minipage}[t]{0.48\linewidth}
\centering
\footnotesize
\setlength{\tabcolsep}{4.0pt}
\renewcommand{\arraystretch}{1.02}
\captionof{table}{\textbf{Pixel statistics.} Four-benchmark averages over executed repairs.}
\label{tab:supp_controls_stats}
\vspace{1mm}
\begin{tabular}{@{}lrrrr@{}}
\toprule
Control & Exec. & Ben. & Harm & Gain \\
\midrule
Cons. score & 107.0 & 1.66M & 0.05M & +0.74 \\
Margin score & 768.3 & 18.51M & 3.44M & +2.64 \\
\rowcolor{SkyBlue!12}
Closed evidence & 1,304.3 & 55.17M & 1.28M & +7.52 \\
\bottomrule
\end{tabular}
\end{minipage}
\vspace{2mm}

\noindent
\begin{minipage}[t]{0.58\linewidth}
\centering
\footnotesize
\setlength{\tabcolsep}{2.7pt}
\captionof{table}{\textbf{Training-free cost.} Selective proposal-level auditing with effective sharded verifier latency.}
\label{tab:supp_efficiency}
\vspace{1mm}
\resizebox{\linewidth}{!}{
\begin{tabular}{lrrrrr}
\toprule
Benchmark & Images & Calls/img & VLM s/img & Prop. s/img & Total s/img \\
\midrule
VOC21 & 1,449 & 1.38 & 0.58 & 0.043 & 0.62 \\
PC60 & 5,104 & 0.39 & 0.16 & 0.043 & 0.21 \\
City & 500 & 4.00 & 1.68 & 0.043 & 1.72 \\
ADE & 2,000 & 1.00 & 0.42 & 0.043 & 0.46 \\
COCO-Object & 5,000 & 0.40 & 0.17 & 0.043 & 0.21 \\
\midrule
Weighted avg. & 14,053 & 0.71 & 0.30 & 0.043 & 0.34 \\
\bottomrule
\end{tabular}}
\end{minipage}
\hfill
\begin{minipage}[t]{0.38\linewidth}
\centering
\footnotesize
\setlength{\tabcolsep}{4.1pt}
\renewcommand{\arraystretch}{1.02}
\captionof{table}{\textbf{Qwen3.5 resources.} Unsharded local single-call profile.}
\label{tab:supp_auditor_resources}
\vspace{1mm}
\begin{tabular}{@{}lrrr@{}}
\toprule
Scale & s/call & Peak & Weights \\
\midrule
0.8B & 0.53 & 2.4 GB & 1.63 GiB \\
4B & 1.79 & 9.4 GB & 8.68 GiB \\
9B & 2.09 & 18.7 GB & 17.98 GiB \\
\bottomrule
\end{tabular}
\end{minipage}
\vspace{2mm}

\noindent
\begin{minipage}[t]{0.48\linewidth}
\centering
\footnotesize
\setlength{\tabcolsep}{4.5pt}
\renewcommand{\arraystretch}{0.98}
\captionof{table}{\textbf{End-to-end timing.} Validation-weighted cost with the default verifier.}
\label{tab:supp_e2e_cost_breakdown}
\vspace{1mm}
\begin{tabular}{lr}
\toprule
Component & Time / image \\
\midrule
Proposal generation & 0.508s \\
Frozen OVSS base forward & 5.042s \\
VLM verifier audit & 0.424s \\
Propagation / writeback & 0.043s \\
\midrule
Cold end-to-end total & 6.018s \\
SEER repair overhead & 0.468s \\
\bottomrule
\end{tabular}
\end{minipage}
\hfill
\begin{minipage}[t]{0.47\linewidth}
\centering
\footnotesize
\setlength{\tabcolsep}{4.6pt}
\renewcommand{\arraystretch}{1.02}
\captionof{table}{\textbf{Cross-base uncertainty.} Bootstrap intervals for the plug-in gains in Table~\ref{tab:supp_cross_base_full}.}
\label{tab:supp_cross_base_ci}
\vspace{1mm}
\begin{tabular}{lcc}
\toprule
OVSS base & $\Delta_{\mathrm{avg}}$ & Bootstrap 95\% CI \\
\midrule
ProxyCLIP & +4.53 & [+4.08, +4.92] \\
ResCLIP & +5.08 & [+4.55, +5.50] \\
Trident & +5.93 & [+5.34, +6.41] \\
CorrCLIP & +5.48 & [+4.86, +5.97] \\
\bottomrule
\end{tabular}
\end{minipage}
\vspace{1mm}

The cost and resource measurements in Tables~\ref{tab:supp_efficiency}, \ref{tab:supp_auditor_resources}, and~\ref{tab:supp_e2e_cost_breakdown} are local reproducibility diagnostics and do not imply real-time inference. The default verifier is a locally deployable open VLM. The 0.42s/call value is the effective latency after sharded execution, while the 2.09s/call resource profile is an unsharded local single-call measurement. Each VLM call audits one proposal using the three-view prompt, and the weighted average is computed over 14,053 validation images. SEER is training-free and does not update OVSS weights; its extra work scales with the number of audited proposals.

\section{Cross-Base Plug-in Diagnostics}
\label{app:cross_base}
The main paper applies the same complete SEER pipeline to multiple OVSS bases. The full-validation per-benchmark results appear in Table~\ref{tab:supp_cross_base_full}, with bootstrap intervals summarized in Table~\ref{tab:supp_cross_base_ci}. Coefficients, thresholds, candidates, verifier, parser, gates, and propagation rules are unchanged; only the base prediction and score cache come from the chosen OVSS method. This controlled plug-in test complements the full eight-benchmark comparison in Tab.~\ref{tab:main_results}.

\begin{table}[H]
\centering
\small
\setlength{\tabcolsep}{3.6pt}
\renewcommand{\arraystretch}{0.94}
\caption{\textbf{Full-validation cross-base plug-in test.} The same SEER rules are used for each OVSS base; $\Delta_{\mathrm{avg}}$ averages four benchmark gains.}
\label{tab:supp_cross_base_full}
\vspace{1mm}
\begin{tabular*}{0.88\linewidth}{@{\extracolsep{\fill}}clcccc@{}}
\toprule
OVSS base & Benchmark & Base & +\method{} & $\Delta$ & $\Delta_{\mathrm{avg}}$ \\
\midrule
\multirow{4}{*}{\makecell[c]{ProxyCLIP}} & VOC21 & 61.3 & 64.9 & +3.6 & \multirow{4}{*}{+4.53} \\
 & PC60 & 35.3 & 39.5 & +4.2 &  \\
 & City & 38.1 & 42.6 & +4.5 &  \\
 & ADE & 20.2 & 26.0 & +5.8 &  \\
\midrule
\multirow{4}{*}{\makecell[c]{ResCLIP}} & VOC21 & 61.1 & 65.2 & +4.1 & \multirow{4}{*}{+5.08} \\
 & PC60 & 33.5 & 38.3 & +4.8 &  \\
 & City & 35.9 & 41.1 & +5.2 &  \\
 & ADE & 18.0 & 24.2 & +6.2 &  \\
\midrule
\multirow{4}{*}{\makecell[c]{Trident}} & VOC21 & 67.1 & 73.0 & +5.9 & \multirow{4}{*}{+5.93} \\
 & PC60 & 38.6 & 44.0 & +5.4 &  \\
 & City & 42.9 & 48.8 & +5.9 &  \\
 & ADE & 21.9 & 28.4 & +6.5 &  \\
\midrule
\multirow{4}{*}{\makecell[c]{CorrCLIP}} & VOC21 & 74.8 & 79.3 & +4.5 & \multirow{4}{*}{+5.48} \\
 & PC60 & 44.2 & 47.7 & +3.5 &  \\
 & City & 49.4 & 55.0 & +5.6 &  \\
 & ADE & 26.9 & 35.2 & +8.3 &  \\
\bottomrule
\end{tabular*}
\vspace{-1mm}
\end{table}

\vspace{-3mm}
\noindent\textbf{Cross-base reading.}
This test is not intended to replace the full eight-benchmark comparison. It asks whether the same deterministic execution contract remains effective when the OVSS prediction and score cache change. All four bases improve on all four benchmarks, with average gains from +4.53 to +5.93 mIoU. This pattern indicates that the repaired residues are not idiosyncratic to CorrCLIP; the evidence-to-execution interface supports plug-in behavior when the new base provides a compatible score field.

\section{Prompt, Policy, and Repair-Space Diagnostics}
\label{app:robustness_policy_upper}
This section audits the repair mechanism from three angles. Prompt and policy sensitivity are reported in Tables~\ref{tab:supp_prompt} and~\ref{tab:supp_policy}; proposal-space headroom is measured with validation-label oracles in Table~\ref{tab:supp_upper_bound}; residual harm and routing behavior are summarized in the remaining diagnostics.

\noindent
\begin{minipage}[t]{0.47\linewidth}
\centering
\footnotesize
\setlength{\tabcolsep}{3.2pt}
\captionof{table}{\textbf{Prompt schemas.} The default prompt has the best average.}
\label{tab:supp_prompt}
\vspace{1mm}
\resizebox{\linewidth}{!}{
\begin{tabular}{lrrrr}
\toprule
Prompt & VOC21 & PC60 & ADE & Avg. \\
\midrule
Concise & +3.75 & +3.26 & +2.58 & +3.20 \\
Neutral & +5.98 & +2.73 & +3.49 & +4.07 \\
\rowcolor{SkyBlue!12}
Default & \textbf{+6.72} & \textbf{+4.19} & \textbf{+9.67} & \textbf{+6.86} \\
\bottomrule
\end{tabular}}
\end{minipage}
\hfill
\begin{minipage}[t]{0.48\linewidth}
\centering
\footnotesize
\setlength{\tabcolsep}{3.2pt}
\captionof{table}{\textbf{Policy sensitivity.} SEER gives the largest gains with broader certified support.}
\label{tab:supp_policy}
\vspace{1mm}
\resizebox{\linewidth}{!}{
\begin{tabular}{lrrrr}
\toprule
Policy & VOC & PC & ADE & Support \\
\midrule
Strict & +3.73 & +2.47 & +7.06 & 0.66M \\
Lenient & +4.56 & +3.91 & +8.23 & 0.82M \\
\rowcolor{SkyBlue!12}
SEER & \textbf{+6.72} & \textbf{+4.19} & \textbf{+9.67} & 1.14M \\
\bottomrule
\end{tabular}}
\end{minipage}
\vspace{1mm}

\noindent\fbox{\begin{minipage}{0.96\linewidth}
\footnotesize
\textbf{Prompt templates.} All prompts use the same legal candidate set $\mathcal{A}_i$ and the same parser; only the amount of visual context and instruction detail changes.
\begin{description}[leftmargin=1.45cm,style=nextline,itemsep=1pt,topsep=1pt]
\item[\textit{Concise.}] ``A region is highlighted in the image. Candidate labels are $\mathcal{A}_i$. Choose exactly one candidate label or \texttt{uncertain}. Return JSON only: \{\texttt{label}, \texttt{confidence}\}.''
\item[\textit{Neutral.}] ``Inspect the highlighted region. The legal labels are $\mathcal{A}_i$; do not use outside labels. Decide whether the region visually matches one candidate or should be \texttt{uncertain}. Return JSON only: \{\texttt{label}, \texttt{confidence}, \texttt{region\_form}\}, where \texttt{region\_form} is one of \texttt{coherent}, \texttt{boundary}, \texttt{mixed}, or \texttt{unclear}.''
\item[\textit{Default.}] ``You are verifying evidence for open-vocabulary semantic segmentation repair. You are shown the global image, a local neighborhood crop, and a tight crop of the same highlighted region. Use global context for scene semantics, local context for surrounding boundaries, and the tight crop for appearance. Candidate labels are $\mathcal{A}_i$ and include \texttt{uncertain}. Do not invent labels. Return JSON only: \{\texttt{label}, \texttt{confidence}, \texttt{region\_form}\}. Use \texttt{uncertain} if the evidence is weak, mixed, outside the candidate set, or not executable as a coherent region.''
\end{description}
\end{minipage}}
\vspace{2mm}

\begin{table}[H]
\centering
\small
\setlength{\tabcolsep}{5pt}
\caption{\textbf{Proposal-space repair headroom.} Oracle rows are analysis-only diagnostics, not inference-time baselines. The full-pool oracle is the headroom row; the top-500 oracle is budget-limited and not a strict upper bound.}
\label{tab:supp_upper_bound}
\vspace{1mm}
\resizebox{0.86\linewidth}{!}{
\begin{tabular}{lrrrr}
\toprule
Benchmark & SEER closed & Oracle top-500 & Oracle full pool & SEER/full \\
\midrule
VOC21 & +6.72 & +13.01 & +16.38 & 0.41 \\
PC60 & +4.19 & +2.33 & +6.56 & 0.64 \\
City & +9.49 & +13.92 & +21.74 & 0.44 \\
ADE & +9.67 & +6.37 & +19.54 & 0.49 \\
\midrule
Average & +7.52 & +8.91 & +16.05 & 0.50 \\
\bottomrule
\end{tabular}}
\vspace{-1mm}
\end{table}

The oracle rows quantify remaining repair headroom inside the same proposal and action space. The full-pool oracle is the strict headroom diagnostic, while the top-500 oracle preserves the audit budget and can be lower than SEER when the budgeted oracle misses actions found by the default replay. Across the four mechanism benchmarks, SEER recovers about half of the full-pool oracle gain on average. COCO-Object is excluded because broad full-pool writes are not a safe upper bound in boundary-sensitive object settings.

\begin{table}[H]
\centering
\footnotesize
\setlength{\tabcolsep}{7pt}
\caption{\textbf{Image/class harm.} Harmed-image rates and losses are measured after SEER repair; negative classes count class-IoU drops.}
\label{tab:supp_harm_distribution}
\vspace{1mm}
\begin{tabular}{lrrrrrr}
\toprule
Benchmark & Images & Harmed img. & Med. loss & P75 loss & Mean gain & Neg. cls. \\
\midrule
VOC21 & 1,449 & 5.8\% & 3.40 & 8.70 & +4.70 & 0 \\
PC60 & 5,104 & 0.9\% & 3.26 & 8.95 & +2.83 & 0 \\
City & 500 & 5.0\% & 0.43 & 1.92 & +6.18 & 0 \\
ADE & 2,000 & 3.8\% & 0.82 & 2.47 & +2.91 & 5 \\
\bottomrule
\end{tabular}
\vspace{-1mm}
\end{table}

\begin{table}[H]
\centering
\footnotesize
\setlength{\tabcolsep}{12pt}
\caption{\textbf{Residual harm sources.} Harmful-pixel rates are measured among modified pixels; source shares are normalized only over residual harmful pixels.}
\label{tab:supp_residual_harm_source}
\vspace{1mm}
\begin{tabular}{lrrr}
\toprule
Benchmark & Harm/mod. & Support spill & Sem. error \\
\midrule
ADE & 2.1\% & 69.8\% & 30.2\% \\
City & 2.1\% & 72.4\% & 27.6\% \\
PC60 & 2.6\% & 57.2\% & 42.8\% \\
VOC21 & 1.4\% & 58.7\% & 41.3\% \\
\bottomrule
\end{tabular}
\vspace{-1mm}
\end{table}

The harm diagnostics address whether average mIoU gains hide systematic damage. Harmed-image rates stay low, no negative classes appear on VOC21, PC60, or City, and ADE has five small negative class deltas, led by stairway $-0.57$, ball $-0.53$, and base $-0.32$. Residual harmful pixels are sparse among modified pixels. Their sources split between support spill, where plausible evidence crosses ambiguous boundaries, and semantic error, where the regional label is wrong. This pattern supports the need for both closed evidence filtering and conservative score-space execution.

\begin{table}[H]
\centering
\small
\setlength{\tabcolsep}{4pt}
\caption{\textbf{Regional-evidence routing taxonomy.} Rates are computed over routed audited proposals. Conservative routing records boundary-aware execution; residual pixel harm is reported in the preceding table.}
\label{tab:supp_failure_taxonomy}
\resizebox{0.98\linewidth}{!}{
\begin{tabular}{lrrrr}
\toprule
Benchmark & Uncertain/low & Candidate miss & Correct high-conf. & Conservative route \\
\midrule
VOC21 & 4.8\% & 0.2\% & 89.6\% & 5.4\% \\
PC60 & 3.7\% & 0.5\% & 87.2\% & 8.6\% \\
City & 3.4\% & 0.3\% & 89.5\% & 6.8\% \\
ADE & 4.6\% & 0.8\% & 84.9\% & 9.7\% \\
Object & 6.9\% & 0.7\% & 81.6\% & 10.8\% \\
\bottomrule
\end{tabular}}
\end{table}

\paragraph{Limitations.}
SEER is a plug-in repair layer over a fixed OVSS prediction, so the complete-system comparison reports the accuracy-cost trade-off of adding frozen proposal-level regional auditing to an OVSS predictor, while matched-input diagnostics isolate repair. Its priority, gate, and routing rules form a deterministic execution contract, fixed across datasets, bases, and auditors to enforce legal labels, high-confidence evidence, and score-compatible support. SEER remains shaped by proposal coverage, candidate recall, regional verifier errors, and latent score support; improving thin structures, mixed proposals, crowded boundaries, and candidate-set misses in boundary-sensitive images is an important direction for future work.

\vspace{1mm}
\section{Visual Examples and Figure Data}
\label{app:visuals}
\begin{figure}[H]
    \centering
    \includegraphics[width=0.98\linewidth]{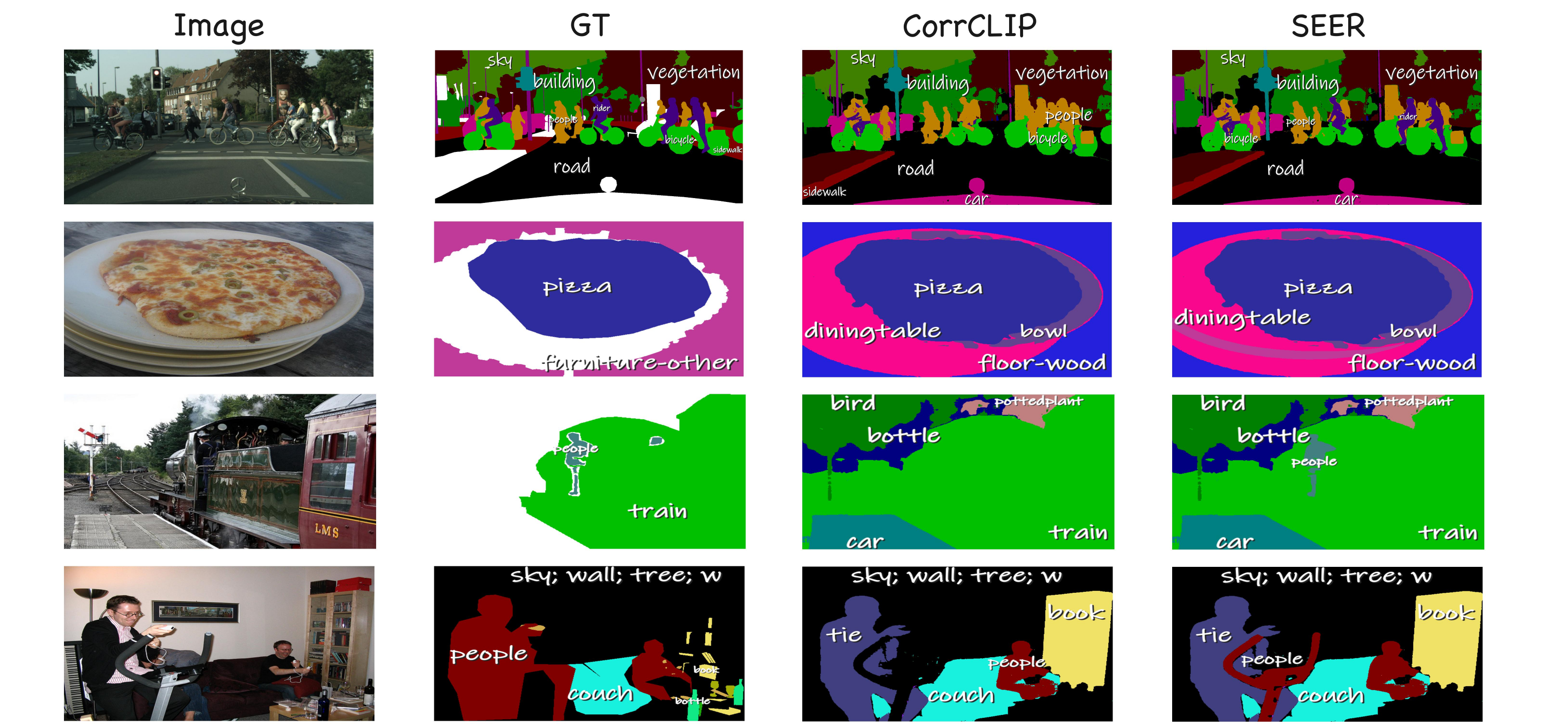}
    \caption{\textbf{Boundary-sensitive and residual failure cases.} SEER can recover proposal-supported regions absent from the initial top-1 prediction, but errors remain when proposals cross semantic boundaries, candidate labels are ambiguous, or the OVSS score field is weak. These cases motivate the execution gate and conservative propagation: plausible regional VLM evidence still needs score-field-constrained pixel support.}
    \label{fig:supp_boundary_sensitive}
\end{figure}

\begin{figure}[p]
    \centering
    \includegraphics[width=0.98\linewidth,height=0.89\textheight,keepaspectratio]{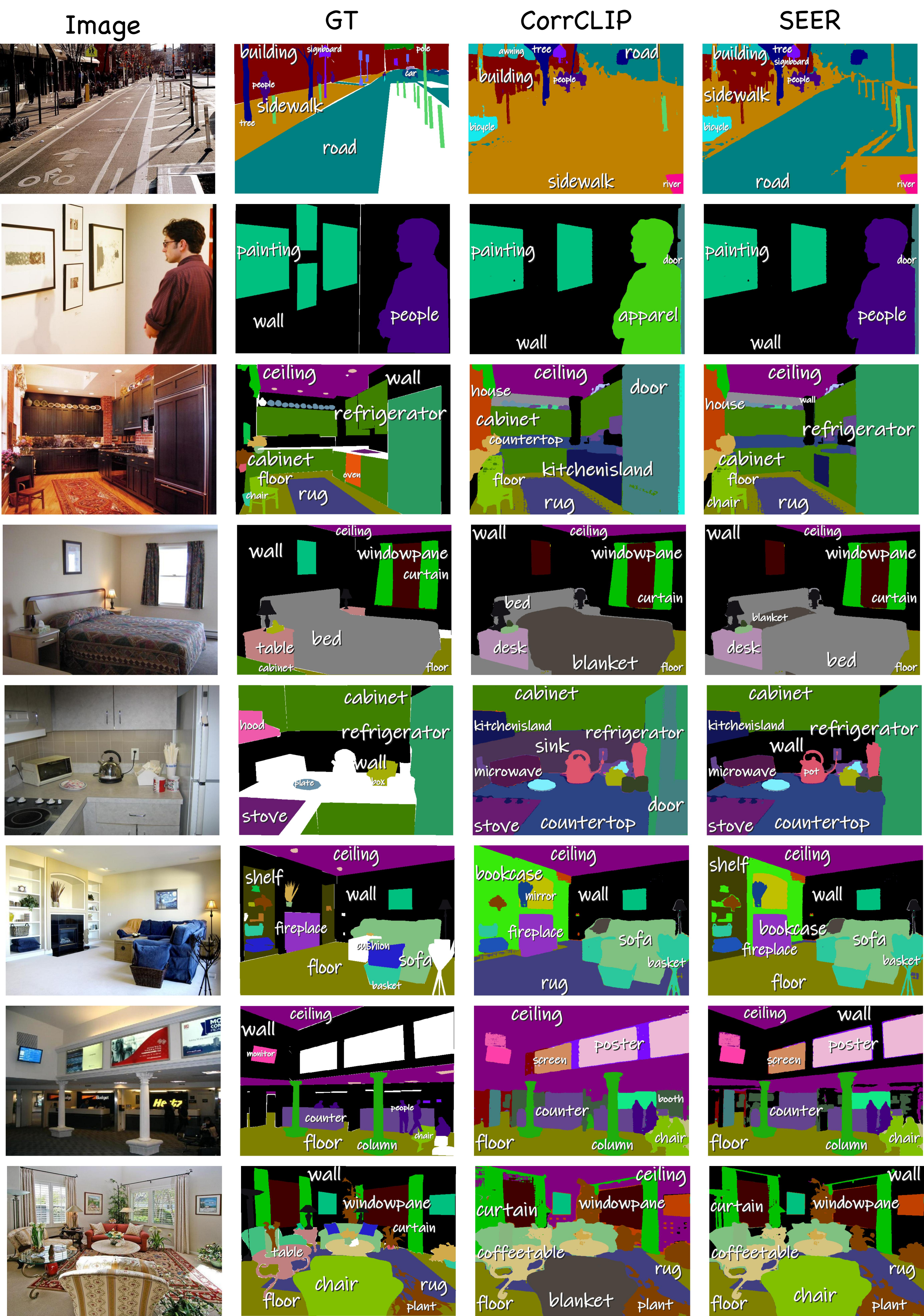}
    \caption{\textbf{ADE20K visual examples.} SEER repairs local semantic residues in indoor and scene-centric images while preserving the surrounding dense prediction structure.}
    \label{fig:supp_visual_ade}
\end{figure}

\begin{figure}[p]
    \centering
    \includegraphics[width=0.98\linewidth,height=0.89\textheight,keepaspectratio]{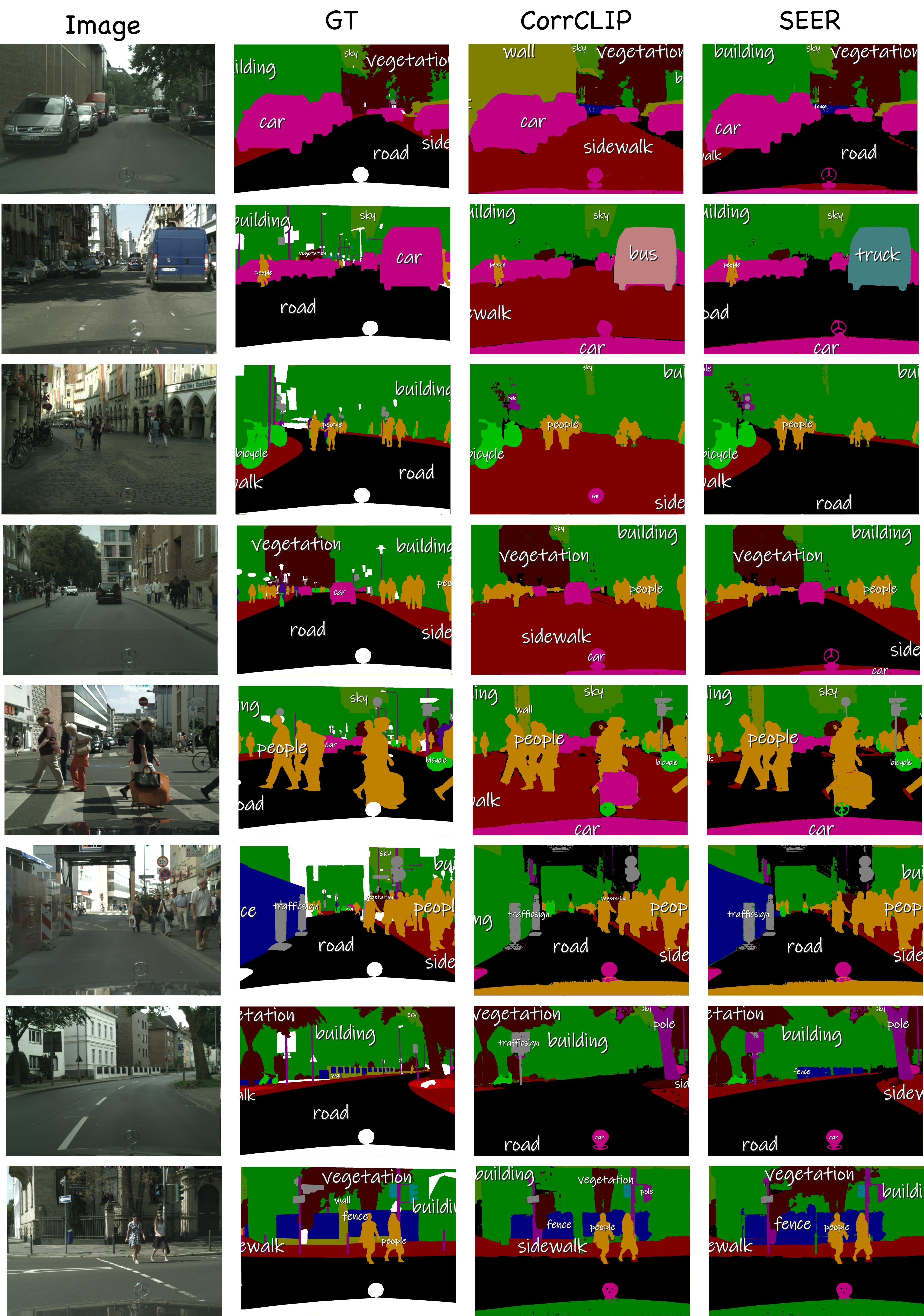}
    \caption{\textbf{Cityscapes visual examples.} SEER targets structured road-scene residues such as missing stuff regions, fragmented objects, and label-space confusions in a fixed OVSS prediction.}
    \label{fig:supp_visual_city}
\end{figure}

\begin{figure}[p]
    \centering
    \includegraphics[width=0.98\linewidth,height=0.89\textheight,keepaspectratio]{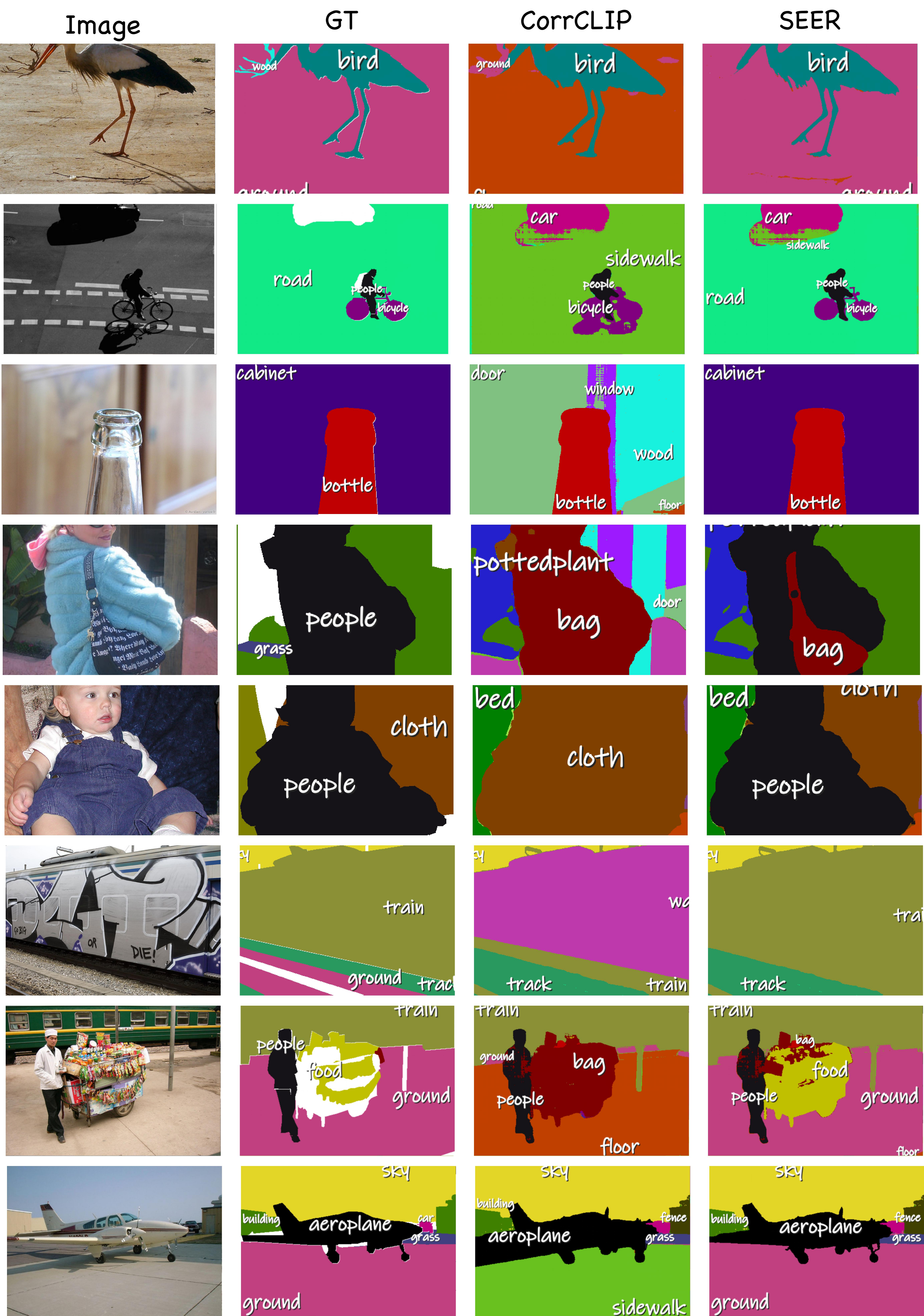}
    \caption{\textbf{PC59 visual examples.} The repairs are localized to semantically supported regions and keep unrelated objects and stuff categories unchanged.}
    \label{fig:supp_visual_ctx59}
\end{figure}

\begin{figure}[p]
    \centering
    \includegraphics[width=0.98\linewidth,height=0.89\textheight,keepaspectratio]{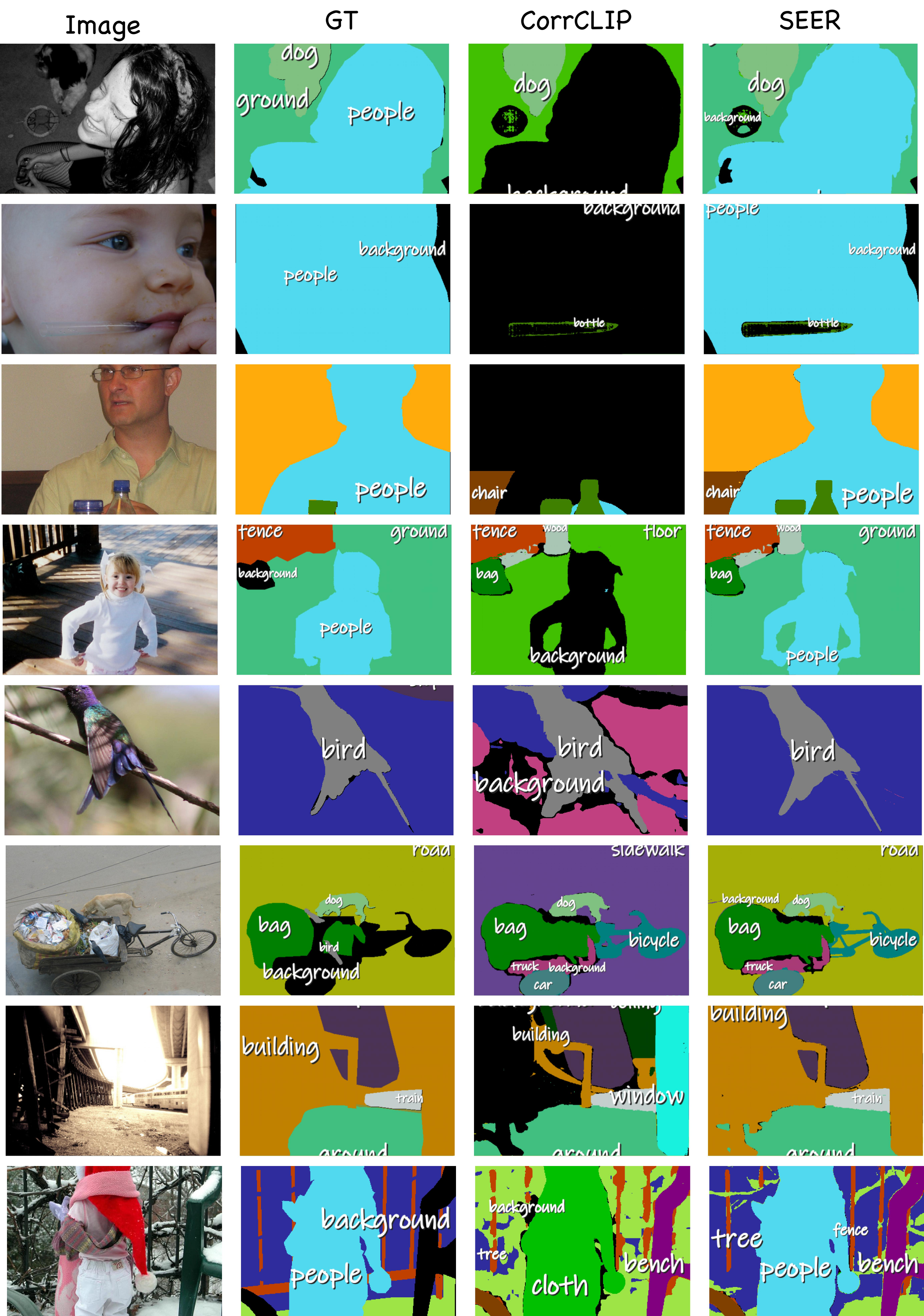}
    \caption{\textbf{PC60 visual examples.} The fixed prediction often contains plausible but incomplete regions; SEER uses closed-set evidence to propagate executable labels where the OVSS score field supports them.}
    \label{fig:supp_visual_ctx60}
\end{figure}

\begin{figure}[p]
    \centering
    \includegraphics[width=0.98\linewidth,height=0.89\textheight,keepaspectratio]{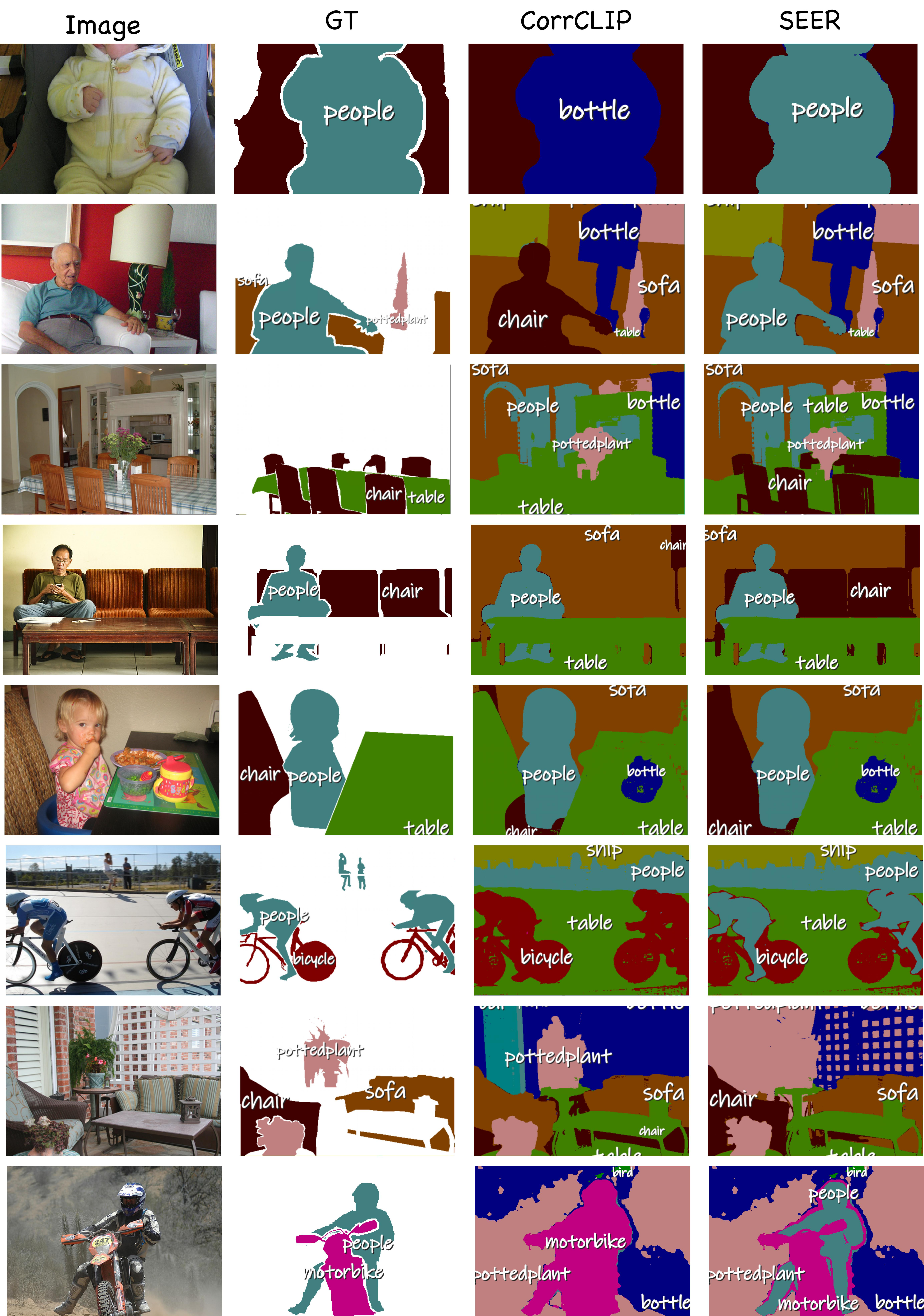}
    \caption{\textbf{VOC20 visual examples.} Object-centric cases show that SEER repairs missing or confused object regions through closed-set evidence and score-space execution.}
    \label{fig:supp_visual_voc20}
\end{figure}

\begin{figure}[p]
    \centering
    \includegraphics[width=0.98\linewidth,height=0.89\textheight,keepaspectratio]{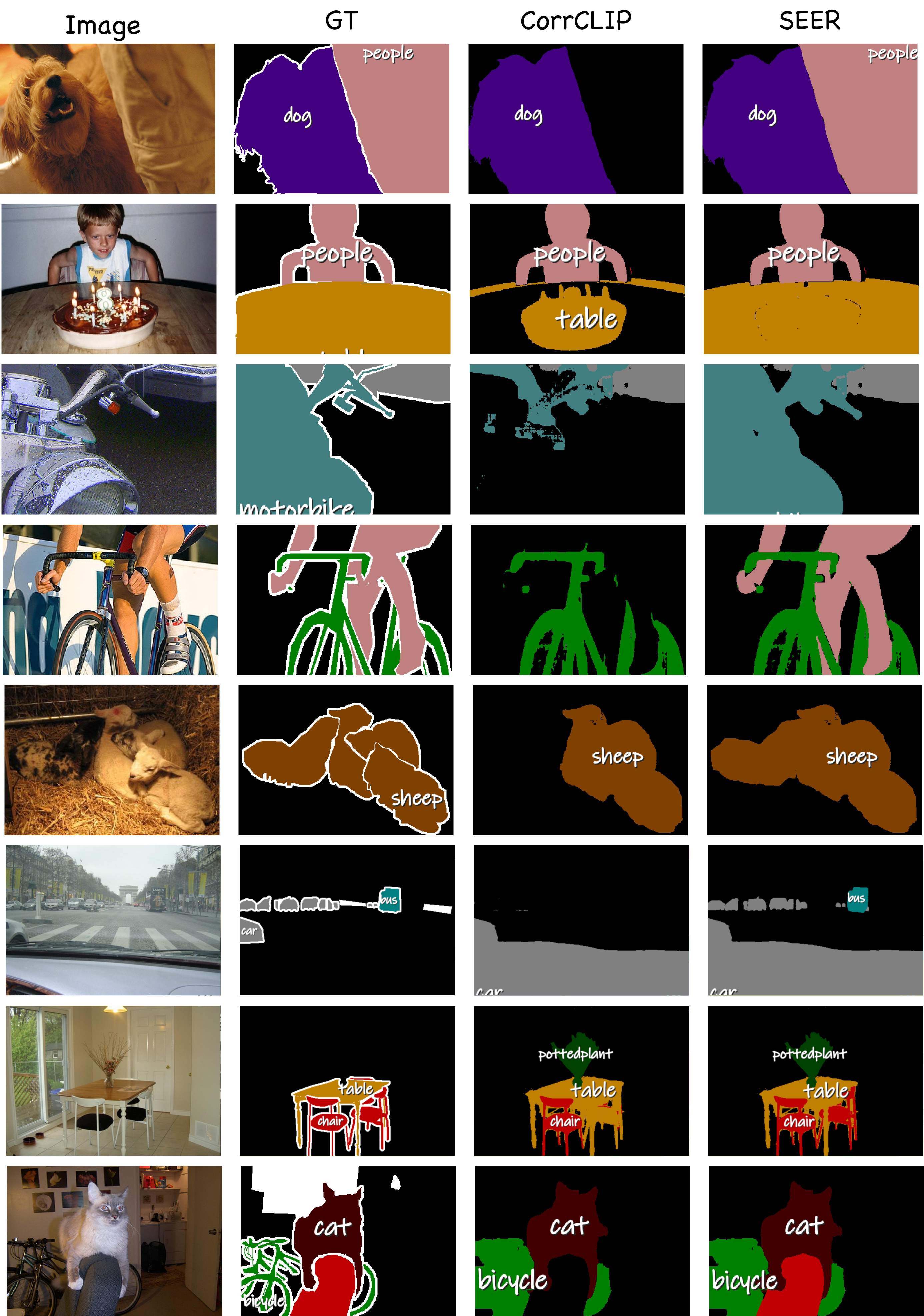}
    \caption{\textbf{VOC21 visual examples.} SEER improves residual object and background errors under the active label space, complementing the quantitative gains reported in the main paper.}
    \label{fig:supp_visual_voc21}
\end{figure}

\begin{figure}[p]
    \centering
    \includegraphics[width=0.98\linewidth,height=0.89\textheight,keepaspectratio]{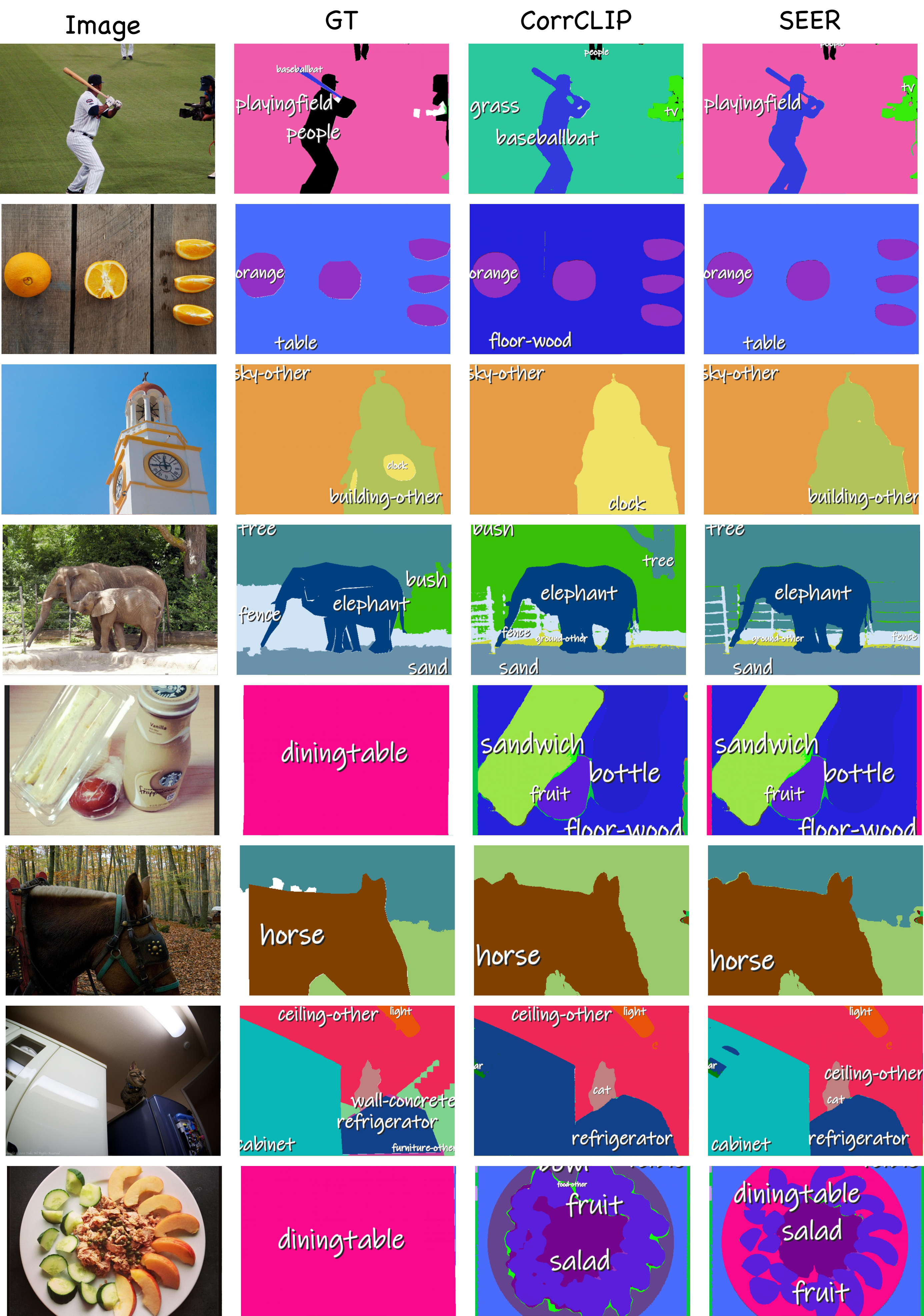}
    \caption{\textbf{COCO-Stuff visual examples.} Stuff-heavy scenes highlight the value of label-space-aware evidence: SEER corrects broad semantic regions while retaining pixel-level execution in the OVSS model.}
    \label{fig:supp_visual_stuff}
\end{figure}

\begin{figure}[p]
    \centering
    \includegraphics[width=0.98\linewidth,height=0.86\textheight,keepaspectratio]{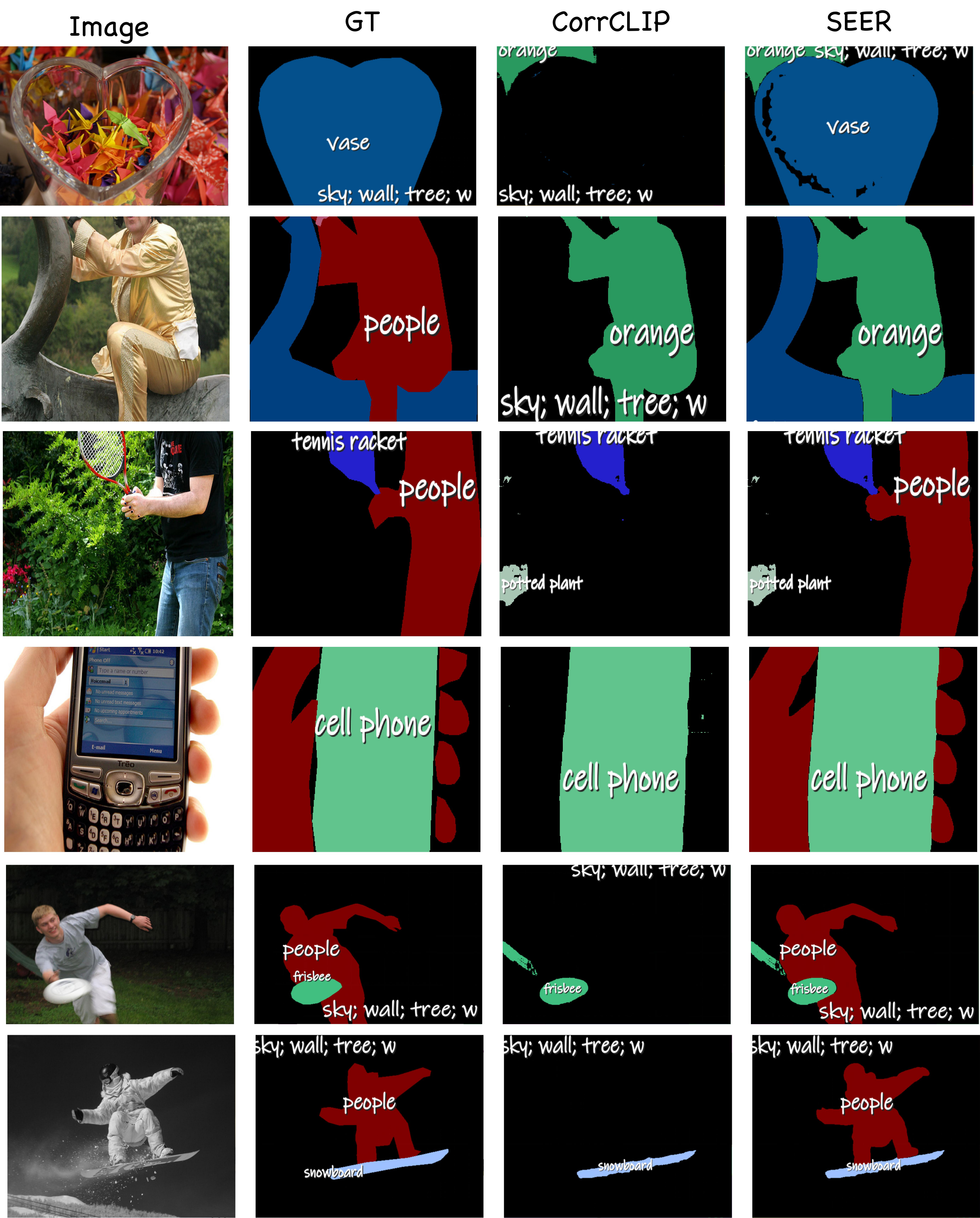}
    \caption{\textbf{COCO-Object visual examples.} Object-boundary cases are more conservative, but SEER still repairs reliable regions when the closed candidate and execution gate agree.}
    \label{fig:supp_visual_object}
\end{figure}

\paragraph{Failure and boundary-sensitive cases.}
Fig.~\ref{fig:supp_boundary_sensitive} shows residual cases where regional recognition is easier than executable support. Crowded street, food/table, and small-object scenes expose four residual risks: over-confident verifier outputs, candidate-set misses, mixed proposals, and weak score-field support. In these cases, a proposal may cover several semantic parts, or the correct label may be visible only in fragmented score support. SEER reduces their impact through abstention and conservative routing, which can leave part of a region unchanged when support is unreliable. Such cases also show the risk of direct VLM pixel authority. Larger galleries in Figs.~\ref{fig:supp_visual_ade}, \ref{fig:supp_visual_city}, \ref{fig:supp_visual_ctx59}, \ref{fig:supp_visual_ctx60}, \ref{fig:supp_visual_voc20}, \ref{fig:supp_visual_voc21}, \ref{fig:supp_visual_stuff}, and~\ref{fig:supp_visual_object} provide held-out examples.

\end{document}